\documentclass{article} %
\usepackage{iclr2022_conference,times}

\usepackage{amsmath,amsfonts,bm}

\def\eqref#1{equation~\ref{#1}}

\def\1{\bm{1}}

\DeclareMathAlphabet{\mathsfit}{\encodingdefault}{\sfdefault}{m}{sl}
\SetMathAlphabet{\mathsfit}{bold}{\encodingdefault}{\sfdefault}{bx}{n}

\usepackage[utf8]{inputenc} %
\usepackage[T1]{fontenc}    %
\usepackage[colorlinks=true,citecolor=black,linkcolor=red]{hyperref}%
\usepackage{url}
\usepackage{booktabs}       %
\usepackage{amsfonts}       %
\usepackage{nicefrac}       %
\usepackage{microtype}      %
\usepackage{xcolor}         %
\usepackage{overpic}
\usepackage{subcaption}

\usepackage{amsmath}
\usepackage{upgreek}
\usepackage{amssymb}
\usepackage{bm}
\usepackage{pifont}%
\newcommand{\cmark}{\ding{51}}%
\newcommand{\xmark}{}%
\usepackage{multirow}
\usepackage{makecell}

\newcommand{\eg}{e.g.}

\usepackage{graphicx,overpic}
\usepackage{tikz}
\usepackage{pgfplots}
\usepackage{wrapfig,lipsum,booktabs}

\pgfplotsset{compat=1.17}
\newlength\savewidth\newcommand\shline{\noalign{\global\savewidth\arrayrulewidth
		\global\arrayrulewidth 1pt}\hline\noalign{\global\arrayrulewidth\savewidth}}

\title{On the Connection between
Local Attention
and Dynamic Depth-wise Convolution}

\newcommand*\samethanks[1][\value{footnote}]{\footnotemark[#1]}

\author{
\centerline{Qi Han$^{1}$\thanks{Equal contribution}~~~~
Zejia Fan$^{2}$\samethanks~~~~
Qi Dai$^{3}$\thanks{Corresponding author.
\href{mailto:wangjingdong@outlook.com}{wangjingdong@outlook.com}}~~~~ Lei Sun$^{3}$~~~~
Ming-Ming Cheng$^{1}$~~~~Jiaying Liu$^{2}$}\\
\centerline{\textbf{Jingdong Wang$^{4}$\samethanks}}\vspace{0.7mm} \\
\centerline{TKLNDST, CS, Nankai Univerisy$^{1}$, Peking University$^{2}$,
Microsoft Research Asia$^{3}$, Baidu Inc.$^{4}$}
}

\iclrfinalcopy %
\begin{document}

\maketitle

\vspace{-0.2cm}
\begin{abstract}
Vision Transformer (ViT)
attains state-of-the-art performance
in visual recognition,
and the variant,
Local Vision Transformer,
makes further improvements.
The major component 
in Local Vision Transformer,
local attention,
performs the attention separately over small local windows.
We rephrase local attention
as a channel-wise locally-connected layer
and analyze it
from two network regularization manners,
sparse connectivity
and weight sharing,
as well as dynamic weight computation.

We point out that local attention resembles depth-wise convolution
and its dynamic variants
in sparse connectivity:
there is no connection across channels,
and each position is connected to 
the positions within a small local window.
The main differences lie in (i) weight sharing
- depth-wise convolution shares connection weights (kernel weights)
across spatial positions
and attention shares the connection weights across channels,
and (ii) dynamic weight computation manners
- local attention is based on dot-products
between pairwise positions in the local window,
and dynamic convolution is based on linear projections
conducted on the center representation
or the globally pooled representation.

The connection {between local attention and dynamic depth-wise convolution}
is empirically 
verified by the ablation study
about weight sharing and dynamic weight computation
in Local Vision Transformer and (dynamic) depth-wise convolution 
based network, namely (dynamic) DWNet. 
We empirically observe that 
the depth-wise convolution based DWNet
and its dynamic variants 
with lower computation complexity
perform on-par with or 
slightly better
than Swin Transformer,
an instance of Local Vision Transformer,
for ImageNet classification, COCO object detection 
and ADE semantic segmentation.
Code is available at \url{https://github.com/Atten4Vis/DemystifyLocalViT}.
\end{abstract}

\vspace{-0.2cm}
\section{Introduction}
\vspace{-0.2cm}
Vision Transformer~\citep{chu2021cpvt,d2021convit,dosovitskiy2021an,guo2021beyond,han2020survey,khan2021transformers,touvron2020training,wang2021pyramid,wu2021cvt,xu2021co,yuan2021tokens} 
has shown promising performance
in ImageNet classification.
The improved variants,
Local Vision Transformer~\citep{chu2021twins,liu2021swin,vaswani2021scaling},
adopt the local attention mechanism,
which partitions the image space into
a set of small windows,
and conducts the attention
over the windows simultaneously.
Local attention leads to great improvement
in memory and computation efficiency
and makes the extension to downstream tasks
easier and more efficient,
such as object detection
and semantic segmentation.

We exploit the network regularization schemes~\citep{goodfellow2016deep},
sparse connectivity that controls the model complexity, 
and weight sharing that relaxes the requirement
of increasing the training data scale 
and reduces the model parameters,
as well as dynamic weight prediction
that increases the model capability,
to study the local attention mechanism. 
We rephrase local attention
as a channel-wise spatially-locally connected layer
with dynamic connection weights.
The main properties are summarized as follows.
(i) Sparse connectivity:
there is no connection across channels,
and each output position
is only connected to the input positions
within a local window.
(ii) Weight sharing:
the connection weights are shared across channels or
within each group of channels.
(iii) Dynamic weight:
the connection weights are dynamically predicted
according to each image instance.

We connect local attention with depth-wise 
convolution~\citep{chollet2017xception, howard2017mobilenets}
and its dynamic variants
that are also a channel-wise spatially-locally connected layer with optional dynamic
connection weights.
They are similar in sparse connectivity.
The main differences lie in (i) weight sharing
- depth-wise convolution shares connection weights (kernel weights)
across spatial positions
and attention shares the connection weights across channels,
and (ii) dynamic weight computation manners
- local attention is based on dot-products
between pairwise positions in the local window,
and dynamic convolution is based on linear projections
conducted on the center representation
or the globally pooled representation.

We further present the empirical verification for the connection.
We take the recently-developed Local Vision Transformer, Swin Transformer~\citep{liu2021swin},
as an example,
and study the empirical performance
of local attention and (dynamic) depth-wise convolution
in the same training settings
as Swin Transformer.
We replace the local attention layer
with the (dynamic) depth-wise convolution layer,
keeping the overall structure unchanged.
The constructed model is named DWNet.

The results show that
the (dynamic) depth-wise convolution-based DWNet achieves comparable or slightly higher performance
for ImageNet classification and two downstream tasks,
COCO object detection and ADE semantic segmentation,
and (dynamic) DWNet takes
lower computation complexity.
The ablation studies imply that 
weight sharing and dynamic weight 
improves the model capability.
Specifically,
(i) for Swin Transformer,
weight sharing across channels is beneficial 
mainly for reducing the 
parameter (attention weight) complexity,
and the attention-based dynamic weight scheme is advantageous 
in learning instance-specific weights and {block-translation equivalent representations};
(ii) for depth-wise convolution,
weight sharing across positions is beneficial for reducing the parameter complexity as well as learning translation equivalent representations,
and the linear projection-based dynamic weight scheme learns instance-specific weights.
\vspace{-.1cm}

\vspace{-0.1cm}
\section{Connecting Local Attention
and Depth-Wise Convolution}
\vspace{-.2cm}
\subsection{Local Attention}
\vspace{-.2cm}
Vision Transformer~\citep{dosovitskiy2021an}
forms a network
by repeating the attention layer and the subsequent point-wise MLP (point-wise convolution).
The local Vision Transformer,
such as Swin Transformer~\citep{liu2021swin} and HaloNet~\citep{vaswani2021scaling},
adopts the local attention layer,
which partitions the space into a set of small windows
and performs the attention operation within each window simultaneously,
to improve the memory and computation efficiency.

The local attention mechanism
forms the keys and values
in a window that the query lies in.
The attention output 
for the query $\mathbf{x}_{i} \in \mathbb{R}^D$ 
at the position $i$ is the aggregation 
of the corresponding values
in the local window,
$\{\mathbf{x}_{i1},
\mathbf{x}_{i2},
\dots,
\mathbf{x}_{iN_k}\}$,
weighted by the corresponding attention weights
$\{a_{i1}, a_{i2}, \dots, a_{iN_k}\}$\footnote{For
presentation convenience,
we ignore
the linear projections conducted
to the queries, 
the keys and the values.
In vision applications,
the value and the corresponding key are from the same
feature
possibly with different linear projections,
and we denote them using the same symbol $\mathbf{x}_{ij}$.}:
\vspace{-0.1cm}
\begin{align}
 \mathbf{y}_i = 
    \sum\nolimits_{j=1}^{N_{k}} a_{ij}\mathbf{x}_{ij},
    \label{eqn:attentionfeature}
    \vspace{-0.1cm}
\end{align}
where $N_k = K_w \times K_h$ is the size of the local window.
The attention weight $a_{ij}$
is computed as 
the softmax normalization of
the dot-product between the query $\mathbf{x}_{i}$ 
and the key $\mathbf{x}_{ij}$:
\vspace{-0.1cm}
\begin{align}
    a_{ij} = \frac{e^{\frac{1}{\sqrt{D}}\mathbf{x}_{i}^\top \mathbf{x}_{ij}}}{Z_i}~~
    \text{where~~} Z_i = \sum\nolimits_{j=1}^{N_{k}} e^{\frac{1}{\sqrt{D}}\mathbf{x}_{i}^\top \mathbf{x}_{ij}}.
    \label{eqn:attentionweight}
    \vspace{-0.1cm}
\end{align}

The multi-head version 
partitions the $D$-dimensional query, key and value vectors
into $M$ subvectors (each with $\frac{D}{M}$ dimensions),
and conducts the attention process 
$M$ times,
each over the corresponding subvector.
The whole output is the concatenation
of $M$ outputs,
$\mathbf{y}_i
= [\mathbf{y}_{i1}^\top~\mathbf{y}_{i2}^\top~
\dots~ \mathbf{y}_{iM}^\top]^\top$.
The $m$th output $\mathbf{y}_{im}$ is calculated by
\vspace{-0.1cm}
\begin{align}
    \mathbf{y}_{im}
    = \sum\nolimits_{j=1}^{N_{k}} a_{ijm}\mathbf{x}_{ijm},
    \label{eqn:mhattentionfeature}
    \vspace{-0.1cm}
\end{align}
where $\mathbf{x}_{ijm}$
is the $m$th value subvector 
and $a_{ijm}$ is the attention weight computed
from the $m$th head
in the same way as Equation~\ref{eqn:attentionweight}.

\begin{figure}
    \centering
    \footnotesize
    (a)~\begin{subfigure}[t]{2.3cm}
       \begin{overpic}[width=0.95\linewidth]{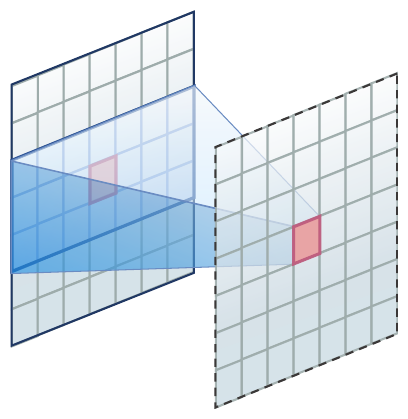}
       \put(-8, 34){\tiny \rotatebox{90}{Spatial}}
       \put(8, 87){\tiny \rotatebox{22}{Channel}}
       \end{overpic}
    \end{subfigure} 
    (b)~\begin{subfigure}[t]{2.3cm}
       \begin{overpic}[width=0.95\linewidth]{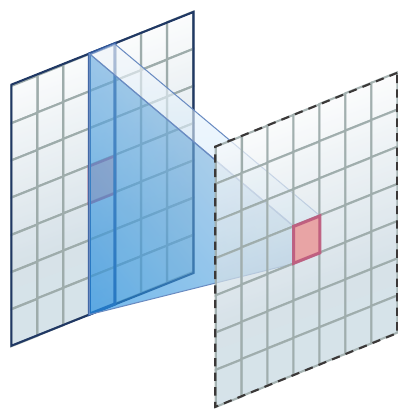}
       \put(-8, 34){\tiny \rotatebox{90}{Spatial}}
       \put(8, 87){\tiny \rotatebox{22}{Channel}}
       \end{overpic}
    \end{subfigure} 
    (c)~\begin{subfigure}[t]{2.3cm}
       \begin{overpic}[width=0.95\linewidth]{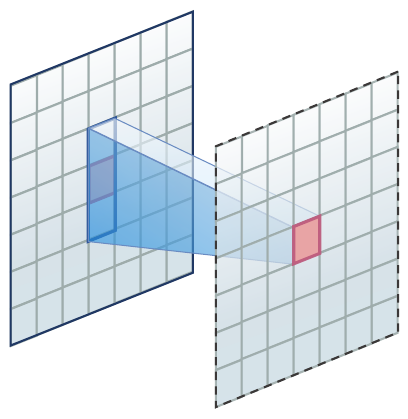}
       \put(-8, 34){\tiny \rotatebox{90}{Spatial}}
       \put(8, 87){\tiny \rotatebox{22}{Channel}}
       \end{overpic}
    \end{subfigure} 
    (d)~\begin{subfigure}[t]{2.3cm}
       \begin{overpic}[width=0.95\linewidth]{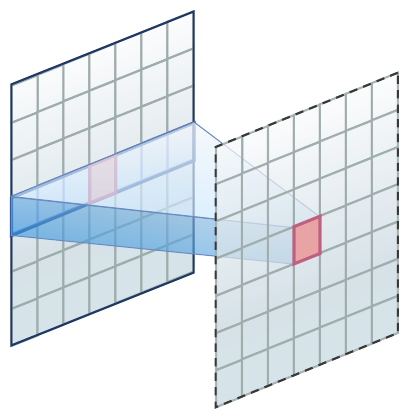}
       \put(-8, 34){\tiny \rotatebox{90}{Spatial}}
       \put(8, 87){\tiny \rotatebox{22}{Channel}}
       \end{overpic}
    \end{subfigure} 
    (e)~\begin{subfigure}[t]{2.3cm}
       \begin{overpic}[width=0.95\linewidth]{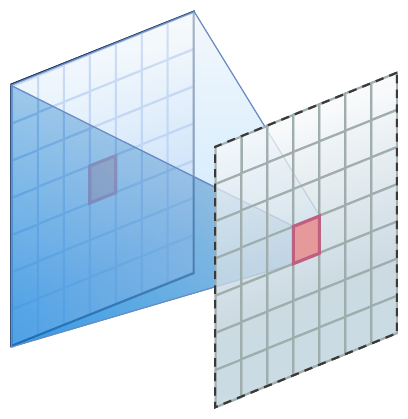}
       \put(-8, 34){\tiny \rotatebox{90}{Spatial}}
       \put(8, 87){\tiny \rotatebox{22}{Channel}}
       \end{overpic}
   \end{subfigure}  
    \caption{\small{ Illustration of
    connectivity} for (a) convolution,
    (b) global attention {and spatial mixing MLP},
    (c) local attention
    and depth-wise convolution,
    (d) point-wise MLP or $1\times 1$ convolution,
    and (e) MLP (fully-connected layer).
    In the spatial dimension,
    we use $1$D to illustrate the local-{connectivity} pattern
    for clarity.
    }
    \vspace{-0.5cm}
\label{fig:connectivity}
\end{figure}
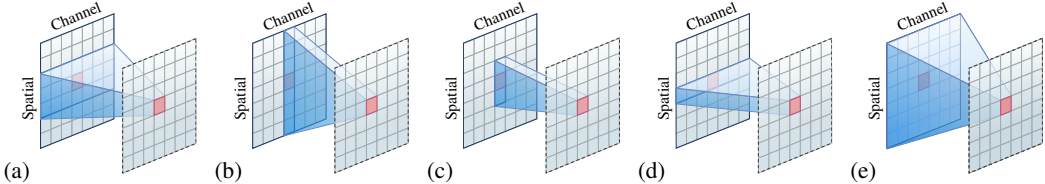

\vspace{-0.1cm}
\subsection{Sparse Connectivity, Weight Sharing, and Dynamic Weight}
\vspace{-.1cm}
We give a brief introduction
of two regularization forms, sparse connectivity and weight sharing,
and dynamic weight,
and their benefits.
We will use the three forms to analyze local attention
and connect it to dynamic depth-wise convolution.

\emph{Sparse connectivity} means
that there are no connections between some output neurons (variables) and some input neurons in a layer.
It reduces the model complexity
without decreasing the number of neurons,
\eg, the size of the (hidden) representations.

\emph{Weight sharing} indicates that 
some connection weights are equal. 
It lowers the number of model parameters
and increases the network size
without requiring a corresponding increase in training data~\citep{goodfellow2016deep}.

\emph{Dynamic weight} refers to learning specialized connection weights for each instance.
It generally aims to increase 
the model capacity.
If regarding the learned connection weights as hidden variables,
dynamic weight can be viewed as introducing
second-order operations
that increase the capability of the network.
The connection to Hopfield networks is discussed in~\citep{Ramsauer20hopfield}.

\vspace{-.1cm}
\subsection{Analyzing local attention}
\vspace{-.1cm}
We show that {local attention} is a {channel-wise spatially-locally connected layer}
with {dynamic weight} computation,
and discuss its properties.
Figure~\ref{fig:connectivity} (c)
illustrates the connectivity pattern.

The aggregation processes~(Equation~\ref{eqn:attentionfeature}
and Equation~\ref{eqn:mhattentionfeature})
for local attention
can be rewritten equivalently in a form
of element-wise multiplication:
\vspace{-0.1cm}
\begin{align}
    \mathbf{y}_i = \sum\nolimits_{j=1}^{N_{k}}
    \mathbf{w}_{ij} \odot \mathbf{x}_{ij},
    \label{eqn:localattentionchannelwiseconv}
    \vspace{-0.1cm}
\end{align}
where
$\odot$ is the element-wise multiplication operator,
and $\mathbf{w}_{ij} \in \mathbb{R}^D$ is the weight vector 
formed from the attention weight $a_{ij}$
or $\{a_{ij1}, a_{ij2}, \dots, a_{ijM}\}$.

\noindent\emph{Sparse connectivity.}
The local attention layer
is spatially sparse:
each position is connected to the $N_k$ positions 
in a small local window.
There are also no connections across channels.
The element-wise multiplication in Equation~\ref{eqn:localattentionchannelwiseconv}
indicates that given the attention weights,
each output element, e.g.,~$y_{id}$
(the $i$th position for the $d$th channel),
is only dependent on
the corresponding input elements
from the same channel
in the window,
$\{x_{i1d}, x_{i2d}, \dots, x_{iN_kd}\}$,
and not related to other channels.

\noindent\emph{Weight sharing.}
The weights are shared with respect to channels.
In the single-head attention case,
all the elements $\{w_{ij1}, w_{ij2},\dots, w_{ijD}\}$
in the weight vector $\mathbf{w}_{ij}$
are the same:
$w_{ijd} = a_{ij}$, $1\leqslant d\leqslant D$.
In the multi-head attention case,
the weight vector $\mathbf{w}_{ij}$
is group-wise same:
$\mathbf{w}_{ij}$ is partitioned
to $M$ subvectors
each corresponding to one attention head, 
$\{\mathbf{w}_{ij1}, \mathbf{w}_{ij2},
\dots, \mathbf{w}_{ijM}\}$,
and the elements in each subvector $\mathbf{w}_{ijm}$ are the same
and are equal to 
the $m$th attention weight,
$a_{ijm}$.

\noindent\emph{Dynamic weight.}
The weights, 
$\{\mathbf{w}_{i1}, \mathbf{w}_{i2},
\dots, \mathbf{w}_{iN_k}\}$,
are dynamically predicted
from the query $\mathbf{x}_i$ and
the keys $\{\mathbf{x}_{i1},
\mathbf{x}_{i2},
\dots,
\mathbf{x}_{iN_k}\}$ 
in the local window
as shown in Equation~\ref{eqn:attentionweight}.
We rewrite it as:
\begin{align}
    \{\mathbf{w}_{i1}, \mathbf{w}_{i2},
\dots, \mathbf{w}_{iN_k}\}
= f(\mathbf{x}_i; \mathbf{x}_{i1}, \mathbf{x}_{i2}, \dots, \mathbf{x}_{iN_k}).
\end{align}
Each weight may obtain the information across all the channels in one head,
and serves as a bridge to deliver the across-channel information
to each output channel.

\noindent\emph{Translation equivalence.}
Different from convolution which satisfies translation equivalence
through sharing weights across positions,
the equivalence to translation
for local attention, 
depends if the keys/values are changed,
i.e., the attention weights are changed,
when the feature map is translated. 

In the case of sparsely-sampled window
(for run-time efficiency), e.g.,~\citep{hu2019local,liu2021swin,ramachandran2019stand,vaswani2021scaling},
local attention is equivalent to block-wise translation, i.e., 
the translation is a block 
or multiple blocks
with the block size same as the window size $K_w \times K_h$, 
and otherwise not equivalent (as keys/values are changed).
In the case that the windows are densely sampled (e.g.,~\citep{Zhao_2020_CVPR}), 
local attention is equivalent to translation.

\noindent\emph{Set representation.}
The keys/values for one query are collected 
as a set
with the spatial-order information lost.
This leads to that
the spatial correspondence between the keys/values
across windows 
is not exploited.
The order information loss
is partially remedied 
by encoding the positions as embeddings~\citep{dosovitskiy2021an,touvron2020training},
or learning a so-called relative position embedding (\eg,~\citep{liu2021swin})
in which the spatial-order information is preserved
as the keys/values in a local window are collected as a vector. 

\subsection{Connection to Dynamic Depth-Wise Convolution}
\label{sec:dynamicdw}
\vspace{-.2cm}
Depth-wise convolution is a type of convolution that applies a single convolutional filter for each channel:
$\bar{\mathbf{X}}_d = \mathbf{C}_d \otimes \mathbf{X}_d$,
where $\mathbf{X}_d$ and $\bar{\mathbf{X}}_d$
are the $d$th input and output channel maps,
$\mathbf{C}_d \in \mathbb{R}^{N_k}$ is the corresponding kernel weight,
and $\otimes$ is the convolution operation.
It can be equivalently written
in the form of element-wise multiplication for each position:
\vspace{-0.1cm}
\begin{align}
\mathbf{y}_{i} = \sum\nolimits_{j=1}^{N_k} 
\mathbf{w}_{\operatornamewithlimits{offset}(i,j)}
\odot
\mathbf{x}_{ij}.
\label{eqn:depthwiseconvolution}
\vspace{-0.1cm}
\end{align}
Here,
$\operatornamewithlimits{offset}(i,j)$ is
the relative offset,
from the $\operatorname{2D}$ coordinate
of the position $j$
to the $\operatorname{2D}$ coordinate of
the central position $i$.
The weights $\{\mathbf{w}_{\operatornamewithlimits{offset}(i,j)}  \in \mathbb{R}^D; j=1,2,\dots,N_k\}$
are reshaped from
$\mathbf{C}_1, \mathbf{C}_2, \dots,
\mathbf{C}_D$.
The $N_k$ weight vectors
are model parameters
and shared for all the positions.

We also consider two dynamic variants
of depth-wise convolution: homogeneous and inhomogeneous\footnote{The homogeneous version follows and applies dynamic convolution to depth-wise convolution.
The inhomogeneous version 
is close to involution~\citep{li2021involution}
and lightweight depth-wise convolution~\citep{wu2019pay}.}.
The homogeneous dynamic variant 
predicts the convolution weights
using linear projections
from a feature vector
that is obtained
by globally-pooling the feature maps:
\begin{align}
    \{\mathbf{w}_1,
    \mathbf{w}_2,
    \dots,
    \mathbf{w}_{N_k}\}
    = g
    (\operatorname{GAP}(\mathbf{x}_1,
    \mathbf{x}_2,
    \dots,
    \mathbf{x}_N)).
\end{align}
Here, $\{\mathbf{x}_1,
    \mathbf{x}_2,
    \dots,
    \mathbf{x}_N\}$
are the image responses.
$\operatorname{GAP}()$ is
the global average pooling operator.
$g()$ is a function based on linear projection:
a linear projection layer to reduce the 
channel dimension with BN and ReLU, 
followed by another linear projection
to generate the connection weights.

The inhomogeneous dynamic variant 
predicts the convolution weights
separately
for each position 
from the feature vector $\mathbf{x}_i$ at the position
(the center of the window):
\begin{align}
    \{\mathbf{w}_{i_1},
    \mathbf{w}_{i_2},
    \dots,
    \mathbf{w}_{i_{N_k}}\}
    = g
    (\mathbf{x}_i).
\end{align}
This means that
the weights are not shared across positions.
We share the weights across the channels
in a way similar to the multi-head attention mechanism
to reduce the complexity.

We describe the similarities and differences
between (dynamic) depth-wise convolution
and local attention.
Figure~\ref{fig:connectivity} (c)
illustrates the connectivity patterns and Table~\ref{tab:regularizationcomparison} shows
the properties between local
attention and depth-wise convolution
, and various other modules.

\noindent\emph{Similarity.}
Depth-wise convolution resembles local attention
in~\emph{sparse connectivity}.
There are no connections across channels.
Each position is only connected to the positions
in a small local window
for each channel.

\noindent\emph{Difference.}
One main difference lies in \emph{weight sharing}:
depth-wise convolution shares the connection weights
across spatial positions,
while local attention shares the weights
across channels
or within each group of channels.
Local attention uses proper weight 
sharing across channels 
to get better performance.
Depth-wise convolution 
benefits from the weight sharing across
positions to reduce the parameter complexity 
and increase the network capability.
 
The second difference is that 
the connection weights for depth-wise convolution
are \emph{static} and
learned as model parameters,
while the connection weights for local attention
are \emph{dynamic}
and predicted from each instance.
The \emph{dynamic} variants
of depth-wise convolution
also benefit from the dynamic weight.

One more difference lies in window representation. 
Local attention represents the positions in a window
by utilizing a \emph{set} form with spatial-order information lost.
It explores the spatial-order information
implicitly
using the positional embedding 
or explicitly using the learned so-called relative positional embedding.
Depth-wise convolution exploits a \emph{vector} form:
aggregate the representations 
within a local window
with the weights indexed by the relative position
(see Equation~\ref{eqn:depthwiseconvolution});
keep spatial correspondence
between the positions for different windows,
thus exploring the spatial-order information
explicitly.

\vspace{-0.1cm}
\section{Experimental Study}
\vspace{-.2cm}
We conduct empirical comparisons
between local attention and depth-wise convolutions 
on three visual recognition tasks:
ImageNet classification,
COCO object detection,
and ADE semantic segmentation.
We follow the structure of Swin Transformer to 
build the depth-wise convolution-based networks.
We apply the same training and evaluation settings from Swin Transformer
to our models. 
In addition,
we study the effects of weight sharing and dynamic weight
in the two structures.
The results for 
large scale pre-training are given in the appendix. 
\vspace{-.2cm}

\begin{table}[t]
    \centering
    \footnotesize
    \setlength{\tabcolsep}{6pt}
        \footnotesize
            \renewcommand{\arraystretch}{1.05}
    \caption{\small{The comparison 
    of attention,
    local MLP (non-dynamic version of local attention,
    the attention weights are learned as static model parameters),
    local attention, 
    convolution,
    depth-wise convolution
    (DW-Conv.) and the dynamic variant (D-DW-Conv.)
    }
    in terms of 
    the patterns of
    sparse connectivity,
    weight sharing, and dynamic weight.
    Please refer to Figure~\ref{fig:connectivity}
    for the connectivity pattern illustration.}
    \label{tab:regularizationcomparison}
    \vspace{-0.3cm}
    \begin{tabular}{l|c c | c |c c | c }
    \shline
         \multirow{2}{*}{} & \multicolumn{2}{c|}{Sparse between positions} &
         Sparse between & \multicolumn{2}{c|}{Weight sharing across}  & Dynamic \\
         \cline{2-3}\cline{5-6}
         & non-local & full & channels & position & channel & weight \\
         \hline
         Local MLP & \cmark & \xmark & \cmark & \xmark & \cmark$^\flat$ & \xmark \\
         Local attention & \cmark & \xmark & \cmark & \xmark & \cmark$^\flat$ & \cmark \\
          DW-Conv. & \cmark & \xmark & \cmark & \cmark & \xmark & \xmark \\
          D-DW-Conv. & \cmark & \xmark & \cmark & \cmark & \xmark & \cmark \\
          Conv. & \cmark & \xmark & \xmark & \cmark  & \xmark & \xmark  \\
         \shline
    \end{tabular}
    \vspace{-0.6cm}
\end{table}
\vspace{-0.1cm}
\subsection{Architectures}
\label{sec:arch}
\vspace{-.2cm}
We use the recently-developed Swin Transformer
as the example of local attention-based networks
and study the performance over the tiny and base networks:
Swin-T and Swin-B,
provided by the authors~\citep{liu2021swin}
We follow the tiny and base networks
to build two depth-wise convolution-based networks,
DWNet-T and DWNet-B,
so that the overall architectures are the same,
making the comparison fair.
We also build two dynamic versions,
dynamic DWNet and i-dynamic DWNet,
by predicting the dynamic weights
as described in Section~\ref{sec:dynamicdw}.
We simply replace local attention in Swin Transformer
by depth-wise convolution of the same window size,
where the pre- and post- linear projections over the values
are replaced by $1\times 1$ convolutions.
We adopt the convolutional network design pattern
to append BN~\citep{ioffe2015batch} and ReLU~\citep{nair2010rectified}
to the convolution.
The details are available in the Appendix. 
In terms of parameter and computation complexity,
the depth-wise convolution-based networks are lower
(Table~\ref{tab:imagenetclassification})
because there are linear projections for keys and values in local attention.
\vspace{-0.1cm}
\subsection{Datasets and Implementation Details} 
\label{sec:details}
\vspace{-.2cm}
\noindent\textbf{ImageNet classification.}
The ImageNet-1K recognition dataset~\citep{deng2009imagenet}
contains $1.28$M training images and $50$K validation images
with totally 1,000 classes.
We use the exactly-same training setting as Swin Transformer~\citep{liu2021swin}. 
The AdamW~\citep{loshchilov2017decoupled} optimizer
for 300 epochs is adopted,
with a cosine decay learning rate
scheduler and 20 epochs of linear warm-up.
The weight decay is $0.05$,
and the initial learning rate is $0.001$. 
The augmentation and regularization 
strategies
include RandAugment~\citep{cubuk2020randaugment}, 
Mixup~\citep{zhang2018mixup}, CutMix~\citep{yun2019cutmix}, 
stochastic depth~\citep{huang2016deep}, etc.

\noindent\textbf{COCO object detection.}
The COCO 2017 dataset~\citep{lin2014microsoft} contains 
$118$K training and $5$K validation images.  
We follow Swin Transformer
to adopt Cascade Mask R-CNN~\citep{cai2019cascade} 
for comparing backbones.
We use the training and test settings from Swin Transformer:
multi-scale training - 
resizing the input such
that the shorter side is between $480$ and $800$ 
and the longer
side is at most $1333$; 
AdamW optimizer with the 
initial learning rate $0.0001$;
weight decay - $0.05$;
batch size - $16$; 
and epochs - $36$.

\noindent\textbf{ADE semantic segmentation.}
The ADE20K~\citep{zhou2017scene} dataset
contains $25$K images, 
$20$K for training, 
$2$K for validation, 
and $3$K for testing,
with $150$ semantic categories.
The same setting as Swin Transformer~\citep{liu2021swin}
is adopted. 
UPerNet~\citep{xiao2018unified} is used
as the segmentation framework.
Details are provided in the Appendix.

\begin{table}[t]
        \centering
            \footnotesize
    \setlength{\tabcolsep}{2.5pt}
            \renewcommand{\arraystretch}{1.14}
            
        \caption{\small ImageNet classification comparison
        for ResNet,
        Mixer and ResMLP,
        ViT and DeiT,
        Swin (Swin Transformer),
        DWNet,
        dynamic DWNet and i-dynamic DWNet.
        }
        \label{tab:imagenetclassification}
        \vspace{-0.3cm}
        \begin{tabular}{l | c r r c | c c }
        \shline 
        method & \makecell{img. size} & \#param. & FLOPs &  \makecell{throughput (img. / s)} & \makecell{top-1 acc.}&  \makecell{real acc.}\\
        \hline 
        \multicolumn{7}{l}{\emph{Bottleneck:
        convolution with low rank}}\\
        \hline
        ResNet-50~\citep{he2016deep} & 224$^2$ & 26M & 4.1G & 1128.3 & 76.2 & 82.5 \\
        ResNet-101~\citep{he2016deep} &  224$^2$ & 45M & 7.9G & 652.0 &  77.4 & 83.7 \\ 
        ResNet-152~\citep{he2016deep} & 224$^2$ &60M & 11.6G & 456.7 & 78.3 & 84.1 \\
        \hline 
        \multicolumn{7}{l}{\emph{Channel and spatial separable MLP, spatial separable MLP = point-wise $1\times 1$ convolution}}\\
        \hline
        
        Mixer-B/16~\citep{tolstikhin2021mlp} & 224$^2$ &46M & - & - & 76.4 & 82.4\\ 
        Mixer-L/16~\citep{tolstikhin2021mlp} & 224$^2$ &189M & - & - & 71.8 & 77.1\\
        ResMLP-12~\citep{touvron2021resmlp} & 224$^2$ &15M & 3.0G & - & 76.6 & 83.3  \\ 
        ResMLP-24~\citep{touvron2021resmlp} & 224$^2$ &30M & 6.0G & - & 79.4 & 85.3\\ 
        ResMLP-36~\citep{touvron2021resmlp} & 224$^2$ &45M & 8.9G & - & 79.7 & 85.6\\ 
       
        \hline 
        \multicolumn{7}{l}{\emph{Global attention: dynamic channel separable MLP + spatial separable MLP}}\\
        \hline
        ViT-B/16~\citep{dosovitskiy2021an} & 384$^2$ &86M & 55.4G & 83.4 & 77.9 & 83.6 \\ 
        ViT-L/16~\citep{dosovitskiy2021an} & 384$^2$ &307M & 190.7G & 26.5 & 76.5 & 82.2 \\
        DeiT-S~\citep{touvron2020training} & 224$^2$ &22M & 4.6G  & 947.3 & 79.8 & 85.7\\
        DeiT-B~\citep{touvron2020training} & 224$^2$ &86M & 17.5G & 298.2 & 81.8 & 86.7\\
        DeiT-B~\citep{touvron2020training} & 384$^2$ &86M & 55.4G & 82.7 & 83.1 & 87.7\\
        \hline 
        \multicolumn{7}{l}{\emph{Local MLP: perform static separable MLP
        in local small windows}}\\
        \hline
        Swin-Local MLP-T & 224$^2$&  26M & 3.8G & 861.0 & 80.3 & 86.1\\
        Swin-Local MLP-B & 224$^2$&  79M & 12.9G & 321.2 & 82.2 & 86.9 \\
        \hline
        \multicolumn{7}{l}{\emph{Local attention: perform attention
        in local small windows}}\\
        \hline
        Swin-T~\citep{liu2021swin} & 224$^2$&28M & 4.5G & 713.5 & 81.3 & 86.6\\
        Swin-B~\citep{liu2021swin} &224$^2$&  88M & 15.4G & 263.0 & 83.3 & 87.9\\
        \hline
        \multicolumn{7}{l}{\emph{Depth-wise convolution + point-wise $1\times 1$ convolution}}\\
        \hline
        DWNet-T &  224$^2$&24M & 3.8G & 928.7 & 81.3 & 86.8 \\
        DWNet-B & 224$^2$& 74M & 12.9G & 327.6 & 83.2 & 87.9\\
        dynamic DWNet-T & 224$^2$ & 51M & 3.8G & 897.0 & 81.9 & 87.3\\
        dynamic DWNet-B & 224$^2$ & 162M & 13.0G & 322.4 & 83.2 & 87.9\\
        i-dynamic DWNet-T & 224$^2$ & 26M & 4.4G & 685.3 & 81.8 & 87.1\\
        i-dynamic DWNet-B & 224$^2$ & 80M & 14.3G & 244.9 & 83.4 & 88.0\\
        \shline
        \end{tabular}
        \vspace{-0.7cm}
\end{table}
\vspace{-0.1cm}
\subsection{Results}
\vspace{-.2cm}
\noindent\textbf{ImageNet classification.}
The comparison for ImageNet classification is given in
Table~\ref{tab:imagenetclassification}.
One can see that
the local attention-based networks,
Swin Transformer,
and the depth-wise convolution-based networks,
DWNets, 
perform on par (with a slight difference of $0.1$) 
in terms of top-$1$ accuracy and real accuracy~\citep{beyer2020we}
for both tiny and base models.
In the tiny model case,
the two dynamic DWNets
perform higher.
In particular,
the depth-wise convolution-based networks
are more efficient 
in parameters and computation complexities.
In the tiny model case,
the parameters and computation complexities are
reduced by $14.2\%$ and $15.5\%$, respectively.
Similarly, in the base model case,
the two costs are
reduced by $15.9\%$ and $16.2\%$, respectively.
The homogeneous dynamic variant takes more parameters
but with almost the same complexity efficiency,
and the inhomogeneous dynamic
variant take advantage of weight sharing across
channels that reduce the model parameters.

\vspace{-0.1cm}
\noindent\textbf{COCO object detection.}  
The comparisons
between local attention (Swin Transformer),
depth-wise convolution (DWNet),
and two versions of dynamic variants
are 
shown in Table~\ref{tab:coco_det_ade_seg}.
Depth-wise convolution performs a little lower than local attention,
and dynamic depth-wise convolution performs better than the static version
and on par with local attention.

\vspace{-0.1cm}
\noindent\textbf{ADE semantic Segmentation.}  
The comparisons of single scale testing
on ADE semantic segmentation
are shown in Table~\ref{tab:coco_det_ade_seg}.
In the tiny model case,
dynamic DWNet is \textasciitilde$1.0\%$
higher than local attention.
In the base model case,
the performances are similar\footnote{
We conducted
an additional experiment 
by changing the ending learning rate from $0$ to $1e-6$.
The base model with depth-wise convolutions
achieves a higher mIoU score: 
$48.9$.}.

\subsection{Empirical Analysis} 
\vspace{-.2cm}
Local and channel-separable connection
has been shown to be helpful
for visual recognition. 
The empirical results in Table~\ref{tab:imagenetclassification},
e.g., local attention performs better than global attention (local connection)
and depth-wise convolution performs better than convolution (channel-separable connection),
also verify it.
In the following,
we present empirical results for weight sharing
and dynamic weight
by taking the tiny models
as examples.

\begin{table}
        \centering
            \footnotesize
    \setlength{\tabcolsep}{2.8pt}
            \renewcommand{\arraystretch}{1.15}
        \caption{\small Comparison results on COCO object detection and ADE semantic segmentation.}
        \label{tab:coco_det_ade_seg}
        \vspace{-0.3cm}
        \begin{tabular}{l | r c| c  c c c  | r r | c}
        \shline
        \multirow{2}{*}{}& \multicolumn{6}{c|}{{COCO Object Detection}} & \multicolumn{3}{c}{{ADE20K Semantic Segmentation}} \\ 
        \cline{2-10}
        
         & \#param. & FLOPs &  AP$^{box}$ & AP$^{box}_{50}$ & AP$^{box}_{75}$ & AP$^{mask}$  & \#param. & FLOPs  &  mIoU  \\
        
        \hline
        Swin-T & 86M & 747G & 50.5 & 69.3 & 54.9 & 43.7  & 60M & 947G &  44.5 \\
        DWNet-T & 82M & 730G & 49.9 & 68.6 & 54.3 & 43.4 & 56M & 928G &   45.5 \\ 
        dynamic DWNet-T & 108M & 730G & 50.5 & 69.5 & 54.6 & 43.7 & 83M & 928G &  45.7 \\ 
        i-dynamic DWNet-T & 84M & 741G & 50.8 & 69.5 & 55.3 & 44.0 & 58M & 939G &  46.2 \\ 
        \hline 
        Swin-B & 145M & 986G & 51.9 & 70.9 & 56.5 & 45.0 & 121M & 1192G & 48.1  \\ 
        DWNet-B & 132M & 924G & 51.1 & 69.6 & 55.4 & 44.2 & 108M & 1129G & 48.3 \\ 
        dynamic DWNet-B & 219M & 924G & 51.2 & 70.0 & 55.4 & 44.4 & 195M & 1129G &  48.0 \\
        i-dynamic DWNet-B & 137M & 948G & 51.8 & 70.3 & 56.1 & 44.8 & 114M & 1153G &  47.8 \\
        \shline
        \end{tabular}
        \vspace{-.4cm}
\end{table}

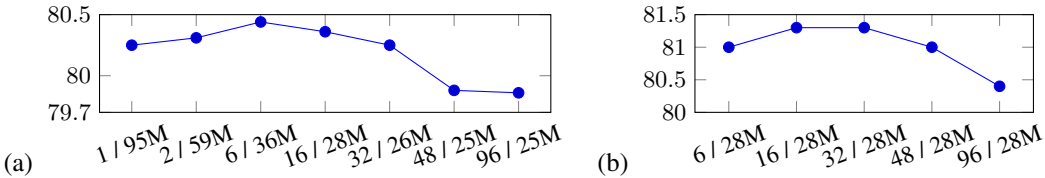
\begin{figure}
    \centering
    (a) \begin{subfigure}[b]{0.5\textwidth}
        \begin{tikzpicture}
        \footnotesize
        \begin{axis}[%
        width=6cm,
        height=1.3cm,
        scale only axis,
        xmin=0.5,
        xmax=7.5,
        xtick={1, 2, 3, 4, 5, 6, 7},
        xticklabel style={rotate=20},
        xticklabels={1 / 95M, 2 / 59M, 6 / 36M, 16 / 28M, 32 / 26M, 48 / 25M, 96 / 25M},
        ymin=79.7,
        ymax=80.5,
        ytick={79.7, 80, 80.5}]

        \addplot+[sharp plot]               
        coordinates                         
        {                               
         (1,80.25) (2,80.31) (3,80.44)
         (4,80.36) (5,80.25) (6,79.88) (7,79.86)
        };
        \end{axis}
        \end{tikzpicture}%
        \vspace{-.1cm}
    \end{subfigure}
    \hfill
    (b) \begin{subfigure}[b]{0.4\textwidth}
        \begin{tikzpicture}
        \footnotesize
        \begin{axis}[%
        width=4.5cm,
        height=1.3cm,
        scale only axis,
        xmin=0.5,
        xmax=5.5,
        xtick={1, 2, 3, 4, 5},
        xticklabels={6 / 28M, 16 / 28M, 32 / 28M, 48 / 28M, 96 / 28M},
        xticklabel style={rotate=20},
        ymin=80.0,
        ymax=81.5,
        ytick={80.0, 80.5, 81.0, 81.5}]
        
        \addplot+[sharp plot]               
        coordinates                         
        {                               
         (5,80.4) (4,81.0) (3,81.3)
         (2,81.3) (1,81.0)
        };
        \end{axis}
        \end{tikzpicture}%
        \vspace{-.1cm}
    \end{subfigure}
    
    \caption{\small 
            Effect of \#channels sharing the weights
            on ImageNet classification.
            X-axis: \#channels within each group / \#param.
            Y-axis: ImageNet classification accuracy.
            (a) Local MLP: the static version of Swin transformer.
            (b) Local attention: Swin transformer.
            Results is reported for tiny
            model on ImageNet dataset.}
        \label{fig:weightsharingacrosschannels}
    \vspace{-0.5cm}
\end{figure}

\noindent\textbf{Weight sharing.}
We study how 
the performance is affected
by
the number of channels in each group across which the weights are shared
(the numbers of attention heads at each stage
are accordingly changed)
for local attention and local MLP (learn the weights in each window
as model parameters 
and not shared across windows).
Figure~\ref{fig:weightsharingacrosschannels} shows the 
effect 
for (a) local MLP - static weights,
and (b) local attention - dynamic weights.
One can see that for local attention,
too many channels and too few channels in each group perform similarly,
but do not lead to the best.
For local MLP, weight sharing significantly reduces
model parameters. These indicate proper weight
sharing across channels is helpful for
both local attention and local MLP.

We further study 
the effect of combining
the weight sharing 
pattern for local MLP and DWNet.
For local MLP, Weight sharing 
across positions means the connection
weight is shared for different spatial blocks in local MLP.
For convolution,
the scheme of sharing weights
across channels
is similar to the multi-head manner
in local attention.
The results in
Table~\ref{tab:positionweighsharing}
suggest that: 
(i)
for local MLP, sharing weight 
across channels reduces the model parameters and sharing across spatial 
blocks do not have big impact;
(ii) For depth-wise convolution,
sharing weight across channels does not 
have big impact, but sharing weight 
across positions significantly increase 
the performance.

The window sampling scheme for 
local MLP and DWNet is 
different:
local MLP sparsely samples the windows
using the way in Swin Transformer,
for reducing the high memory cost, 
and DWNet densely samples the windows.
Weight sharing across positions in local MLP is 
insufficient for learning translation-equivalent 
representation,
explaining why local MLP with weight sharing
across both channels and positions
performs lower than
depth-wise convolution
with additional weight sharing across channels.

\begin{table}[t!]
\centering
    \footnotesize
    \setlength{\tabcolsep}{9pt}
    \renewcommand{\arraystretch}{1.12}
    \caption{
    Effects of weight sharing across channels and positions.
    The results are reported on the ImageNet top-1 accuracy.
    SC = Sharing across channels. SP = sharing across positions.}
    \label{tab:positionweighsharing}
    \vspace{-0.3cm}
    \begin{tabular}{c | c c   c c | c|  c c   c c}
    \shline
      & SC & SP & Acc. & \#param. & & SC & SP & Acc. & \#param. \\
      \hline
    \multirow{3}{*}{Local MLP} & \ding{55} & \cmark & 80.2 & 35.3M & \multirow{3}{*}{DWNet}    & \ding{55} & \cmark & 81.3 & 24.2M \\
                           & \cmark & \ding{55} & 80.3 & 26.2M & & \cmark & \ding{55} & 80.3 & 26.2M\\
                           & \cmark & \cmark & 80.3 & 24.3M  & & \cmark & \cmark & 81.1 & 23.9M \\
    \shline
    \end{tabular}
    \vspace{-0.4cm}
\end{table}

\noindent\textbf{Dynamic weight.} 
We study how dynamic weight in local attention
affects performance.
As seen from Table~\ref{tab:imagenetclassification},
local MLP achieves, the static version, $80.3\%$ and $82.2\%$ for tiny and base models,
lower than Swin, the dynamic version, $81.3\%$ 
and $83.3\%$.
This implies that dynamic weight is helpful.
The improvements from dynamic weight
are also observed for depth-wise convolution 
(Table~\ref{tab:imagenetclassification}).

We further study the effects
of the attention scheme and the linear-projection scheme
for dynamic weight computation.
The observations from in Table~\ref{tab:different_dynamic}
include:
the attention mechanism for shifted and sliding window sampling
performs similarly;
the inhomogeneous dynamic weight computation way is better 
than the attention mechanism ($81.8$ vs $81.4$).
We think that the reasons for the latter observation
include:  
for the attention mechanism the representation is only
block translation equivalent other than
any translation equivalent;
the linear projection-based dynamic weight scheme (vector representation for the window)
learns better weights than the attention-based scheme
(set representation for the window).
We also observe that
such influence
is eliminated for large models and detection tasks.

\noindent\textbf{Set representation.}
Local attention represents the positions in a window
as a set
with the spatial-order information lost.
Swin Transformer learns relative positional embeddings 
where the positions in a window are
actually described as a vector keeping the spatial-order information.
It is reported in~\citep{liu2021swin}
that removing the relative positional embeddings
leads to a $1.2\%$ accuracy drop,
indicating the spatial-order information is important.

\begin{table}[t]
    \centering
    \footnotesize
    \setlength{\tabcolsep}{3pt}
    \renewcommand{\arraystretch}{1.15}
    \caption{
    \small{
    Comparison of different dynamic weight manners.
    The results are reported on the ImageNet top-1 accuracy.
    Shifted window sampling (Win. samp.) means the way in Swin Transformer
    and sliding means the densely-sampling manner.
    The result of Sliding local MLP is from~\citep{liu2021swin}.
    homo. dyna. = homogeneous dynamic weight.
    inhomo. dyna. = inhomogeneous dynamic weight.
    }}
    \label{tab:different_dynamic}
    \vspace{-0.3cm}
    \begin{tabular}{c | c c  c  c | c | c c  c c}
    \shline
      & Win. samp.  & \#param. & FLOPs & Acc. & & Win. samp.  & \#param. & FLOPs & Acc. \\
    \hline
    Local MLP   & shifted   & 26M & 3.8G & 80.3 & DWNet & sliding  & 
    24M & 3.8G & 81.3\\
    w/ attention & shifted & 28M & 4.5G & 81.3 & w/ homo. dyna.  & sliding  &  51M & 3.8G & 81.9\\
    w/ attention & sliding   & 28M & 4.5G & 81.4 & w/ inhomo. dyna. & sliding  &  26M & 4.4G & 81.8\\
    \shline
    \end{tabular}
    \vspace{-0.2cm}
\end{table}

\begin{table}[t]
    \centering
    \footnotesize
    \setlength{\tabcolsep}{20pt}
    \renewcommand{\arraystretch}{1.15}
    \caption{Comparison with concurrent works on ImageNet classification with tiny models. 
    }
    \label{tab:additionaldw}
    \vspace{-0.3cm}
    \begin{tabular}{l | c c c c }
    \shline
      & \#param. & FLOPs & top-1 acc.\\
      \hline
      Twins-PCPVT~\citep{chu2021twins} & 24M & 3.8G & 81.2 \\
      Twins-SVT~\citep{chu2021twins} & 24M & 2.9G & 81.7\\
      CoaT-Lite~\citep{xu2021co} & 20M & 4.0G & 81.9 \\
      CoaT~\citep{xu2021co} & 22M & 12.6G & 82.1 \\
      PVT-v2~\citep{wang2021pvtv2} & 25M & 4.0G & 82.0 \\ 
      Shuffle Transformer~\citep{huang2021shuffle} & 29M & 4.6G & 82.5 \\
      \hline 
      i-dynamic DWNet & 26M & 4.4G & 81.8 \\
      i-dynamic DWNet + DW & 27M & 4.4G & 82.3 \\
    \shline
    \end{tabular}
    \vspace{-0.4cm}
\end{table}

\noindent\textbf{Concurrent works.}
We give the comparison 
between inhomogeneous dynamic depth-wise convolution (i-dynamic DWNet)
and concurrent local attention-based works~\citep{chu2021twins,wang2021pvtv2,huang2021shuffle,xu2021co}
in Table~\ref{tab:additionaldw}.
We follow Shuffle Transformer
and add an extra DW Conv. before FFN in i-dynamic DWNet,
and the performance is improved by $0.5$.
The performance is on par with 
these concurrent works except the 
Twins-SVT (81.9\%, 2.9G) which uses interleaved attention
and additional depth-wise convolutions.

\section{Related Work}
\vspace{-.3cm}

\noindent\textbf{Sparse connectivity.}
{Sparse connection across channels}
is widely explored  
for removing redundancy in the channel domain.
The typical schemes are depth-wise convolution
adopted by MobileNet~\citep{howard2017mobilenets,sandler2018mobilenetv2},
ShuffleNetV2~\citep{ma2018shufflenet}
and
IGCv3~\citep{sun2018igcv3},
and group convolution
adopted by ResNeXt~\citep{xie2017aggregated}, 
merge-and-run~\citep{ZhaoLMLZZTW18},
ShuffleNetV1~\citep{zhang2018shufflenet}, 
and IGC~\citep{zhang2017interleaved}.

The self-attention unit\footnote{The pre- and post-
linear projections for values can be regarded as $1\times 1$
convolutions.
The attention weights generated from keys and values
with linear projections 
in some sense mix the information across channels.} 
in Vision Transformer,
its variants~\citep{chen2020pre,chu2021cpvt,dosovitskiy2021an,han2021transformer,heo2021rethinking,li2021involution,liu2021swin,pan2021scalable,touvron2020training,vaswani2021scaling,wang2021pyramid,wu2021cvt,yuan2021incorporating,yuan2021tokens,zhang2021multi,Zhao_2020_CVPR,zhou2021deepvit},
and the spatial information fusion unit
(\eg, token-mixer in MLP-Mixer~\citep{tolstikhin2021mlp} and
ResMLP~\citep{touvron2021resmlp})
have no connections
across channels.

$1\times 1$ 
(point-wise) convolution
(in ShuffleNetV2~\citep{ma2018shufflenet}, MobileNet~\citep{howard2017mobilenets,sandler2018mobilenetv2}, 
IGC~\citep{zhang2017interleaved}, ViT~\citep{dosovitskiy2021an},
local ViT~\citep{liu2021swin,vaswani2021scaling}, MLP-Mixer~\citep{tolstikhin2021mlp}, ResMLP~\citep{touvron2021resmlp})
has no connections across spatial positions.
The convolutions with other kernel sizes
and local attention~\citep{Zhao_2020_CVPR, liu2021swin,vaswani2021scaling} 
have connections between each position and the positions within a small local window, respectively.

\noindent\textbf{Weight sharing.}
{Weight sharing
across spatial positions} is mainly used 
in convolution, including normal convolution,
depth-wise convolution and point-wise convolution.
{Weight sharing across channels} is adopted in the attention
unit~\citep{vaswani2017attention},
its variants~\citep{chu2021twins,chu2021cpvt,dosovitskiy2021an,li2021involution,liu2021swin,touvron2020training,vaswani2021scaling,wang2021pyramid,wu2021cvt,yuan2021tokens},
and token-mixer MLP in MLP-mixer~\citep{tolstikhin2021mlp} and ResMLP~\citep{touvron2021resmlp}.

\noindent\textbf{Dynamic weight.}
Predicting the connection weights
is widely studied in convolutional networks.
There are basically two types.
One is to learn homogeneous connection weights,
\eg, 
SENet~\citep{hu2018squeeze}, dynamic convolution~\citep{jia2016dynamic}.
The other is to learn the weights for each region or each position (GENet~\citep{hu2018gather}, Lite-HRNet~\citep{yu2021lite}, Involution~\citep{li2021involution}).
The attention unit in ViT or local ViT
learns dynamic connection weights for each position.

\noindent\textbf{Networks built
with depth-wise separable convolutions.}
There are many networks built upon depth-wise separable convolution
or its variants,
such as MobileNet~\citep{howard2017mobilenets,sandler2018mobilenetv2},
ShuffleNet~\citep{ma2018shufflenet}, IGC~\citep{zhang2017interleaved},
Xception~\citep{chollet2017xception},
and
EfficientNet~\citep{tan2019efficientnet,tan2021efficientnetv2}.
In this paper,
our goal is to connect dynamic depth-wise convolution with local attention.

\noindent\textbf{Convolution vs Transformer.}
The study in~\citep{CordonnierLJ20}
shows that a multi-head self-attention
layer
can simulate a convolutional layer
by developing additional carefully-designed 
relative positional embeddings
with~\emph{the attention part dropped}.
Differently,
we connect (dynamic) depth-wise convolution and local
self-attention by~\emph{connecting the attention weights for
self-attention and the dynamic weights for convolution}
(as well as studying weight sharing).
In~\citep{Jean19},
the mathematical connection (in terms of the tensor form) 
between convolution and attention 
is presented.
The opinion 
that convolution and attention
are essentially about the model complexity control
is similar to ours,
and we make the detailed analysis 
and report empirical studies.

The concurrently-developed work in NLP~\citep{tay2021pretrained}
empirically compares lightweight depth-wise convolution~\citep{wu2019pay}
to Transformer for NLP tasks,
and 
reaches a conclusion similar to ours
for vision tasks:
convolution and Transformer obtain on-par results.
Differently,
we attempt to understand
why they perform on par
from three perspectives:
sparse connectivity,
weight sharing 
and dynamic weight,
and discuss their similarities and differences.

\vspace{-.3cm}
\section{Conclusion}
\vspace{-.3cm}
The connections
between local attention and dynamic depth-wise convolution
are summarized as follows.
(i) Same with dynamic depth-wise convolution,
local attention benefits from
two sparse connectivity forms:
local connection
and no connection across channels.
(ii) Weight sharing across channels in local attention
is helpful for reducing 
the parameter (attention weight) complexity
and slightly boosting the performance,
and weight sharing across positions in depth-wise convolution
is helpful for reducing the parameter complexity
and learning translation-equivalent representations and thus boosting the performance.
(iii) The attention-based dynamic weight computation for local attention
is beneficial 
for learning image-dependent weights
and block-translation equivalent representations,
and the linear projection-based 
dynamic weight computation for (in)homogeneous dynamic depth-wise convolution is beneficial 
for learning image-dependent weights.
The constructed i-dynamic DWNet
is superior over Swin transformer
for ImageNet classification and segmentation in the case of tiny models,
and on par for larger models and detection tasks.
In addition,
the better downstream performance 
for local attention and depth-wise convolution 
stems from the larger kernel size
($7\times 7$ vs $3\times 3$),
which is also observed in~\cite{yuan2021hrformer}.

\clearpage
\bibliographystyle{iclr2022_conference}

\clearpage
\appendix
{\LARGE \textsc{Appendix}}
\section{Relation Graph}
\begin{figure}[h]
    \centering
    \includegraphics[width=0.8\linewidth]{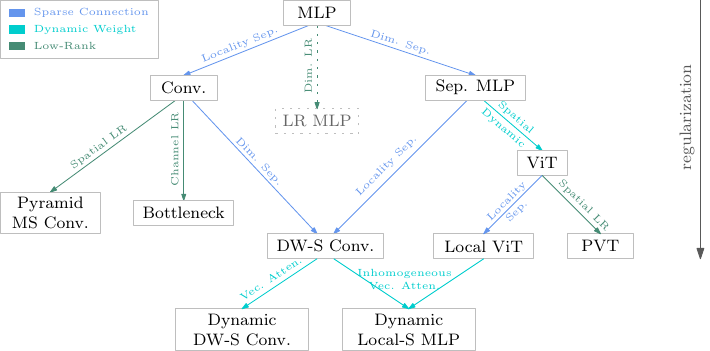}
    \caption{\small{
    Relation graph} for convolution (Conv.),
    depth-wise separable convolution (DW-S Conv.), Vision Transformer (ViT) building block,
    local ViT building block,
    Sep. MLP (\eg,~MLP-Mixer and ResMLP),
    dynamic depth-wise separable convolution (Dynamic DW-S Conv.),
    as well as dynamic local separable MLP (
    e.g., involution~\citep{li2021involution} and inhomogeneous dynamic depth-wise convolution)
    in terms of sparse connectivity
    and
    dynamic weight. 
    Dim. = dimension including spatial and channel,
    Sep. = separable,
    LR = low rank,
    MS Conv. = multi-scale convolution,
    PVT = pyramid vision transformer.
    }
\label{fig:relation1}
\end{figure}

We present a relation graph in Figure~\ref{fig:relation1} 
to describe the relation between convolution, 
depth-wise separable convolution (depth-wise convolution + 
$1\times1$ convolution), Vision Transformer, Local Vision
Transformer, as well as multilayer perceptron (MLP),
Separable MLP in terms of 
sparse connectivity, weight sharing,
and dynamic weight.
Table~\ref{tab:regularizationcomparisonappendix}

Multilayer perceptron (MLP)
is
a fully-connected layer: 
each neuron (an element
at each position and each channel) 
in one layer is connected with all the neurons
in the previous layer\footnote{We use the widely-used definition
for the term MLP: fully-connected layer.
There might be other definitions.}.
Convolution and separable MLP are sparse versions of MLP.
The connection weights can be formulated as a tensor
(e.g., $3$D tensor, two dimension for space and one dimension for channel)
and the low-rank approximation of the tensor can be used to
regularize the MLP.

Convolution is a locally-connected layer,
formed
by connecting each neuron to the neurons in a small local window
with the weights shared across the spatial positions.
Depth-wise separable convolution
is formed by decomposing the convolution 
into two components:
one is point-wise $1\times 1$ convolution, mixing the information across channels,
and the other is depth-wise convolution, mixing the spatial information.
Other variants of convolution, 
such as bottleneck, multi-scale convolution or pyramid,
can be regarded as low-rank variants.

Separable MLP (e.g., MLP-Mixer and ResMLP)
reshapes the $3$D tensor
into a $2$D format
with the spatial dimension
and channel dimension.
Separable MLP consists of two sparse MLP
along the two dimensions separately,
which are formed 
by separating the input neurons
into groups.
Regarding channel sparsity,
the neurons in the same channel form a group,
and an MLP is performed over each group 
with the MLP parameters shared across groups,
forming the first sparse MLP (spatial/token mixing).
A similar process is done by viewing the neurons at the same position into a group,
forming the second sparse MLP (channel mixing).

Vision Transformer is a dynamic version of separable MLP.
The weights in the first sparse MLP (spatial/token mixing)
are dynamically predicted
from each instance.
Local Vision Transformer is a spatially-sparser version of Vision Transformer:
each output neuron is connected 
to the input neurons in a local window.
PVT~\citep{wang2021pyramid} is a pyramid (spatial sampling/ low-rank) variant of Vision Transformer.

Depth-wise separable convolution can also be regarded
as a spatially-sparser version of separable MLP.
In the first sparse MLP (spatial/token mixing),
each output neuron is only dependent on
the input neurons in a local window,
forming depth-wise convolution.
In addition,
the connection weights are shared across spatial positions, instead of across channels.

\section{Matrix Form Explanation}
\label{sec:matrixformrelationgraph}
We use the matrix form to explain sparsity connectivity in various layers
and how they are obtained
by modifying the MLP.

\noindent\textbf{MLP.}
The term MLP, Multilayer Perceptron, is used ambiguously, sometimes loosely to any feedforward neural network.
We adopt one of the common definitions,
and use it to refer to 
fully-connected layers.
Our discussion
is based on a single fully-connected layer,
and can be easily generalized
to two or more fully-connected layers.
One major component,
except the nonlinear units and others,
is a linear transformation:
\begin{align}
    \mathbf{y} = \mathbf{W}\mathbf{x},
    \label{eqn:MLP}
\end{align}
where $\mathbf{x}$ represents the input neurons,
$\mathbf{y}$ represents the output neurons,
and $\mathbf{W}$ represents the connection weights,
e.g., $\mathbf{W} \in \mathbb{R}^{NC \times NC}$,
where $N$ is the number of positions,
and $C$ is the number of channels.

\begin{table}[t]
    \centering
    \footnotesize
    \setlength{\tabcolsep}{6pt}
        \footnotesize
            \renewcommand{\arraystretch}{1.05}
    \caption{\small{The comparison 
    of attention,
    local MLP (non-dynamic version of local attention,
    the attention weights are learned as static model parameters),
    local attention, 
    convolution,
    depth-wise convolution
    (DW-Conv.) and the dynamic variant (D-DW-Conv.),
    as well as MLP and MLP variants}
    in terms of 
    the patterns of
    sparse connectivity,
    weight sharing, and dynamic weight.
    $^\dagger$Spatial-mixing MLP (channel-separable MLP) corresponds to token-mixer MLP. 
    $^\ddagger 1\times 1$ Conv. is also called point-wise (spatial-separable) MLP. 
    $^\flat$The weights might be shared within each group of channels.
    Please refer to Figure~\ref{fig:connectivity}
    for the connectivity pattern illustration.}
    \label{tab:regularizationcomparisonappendix}
    \begin{tabular}{l|c c | c |c c | c }
    \shline
         \multirow{2}{*}{} & \multicolumn{2}{c|}{Sparse between positions} &
         Sparse between & \multicolumn{2}{c|}{Weight sharing across}  & Dynamic \\
         \cline{2-3}\cline{5-6}
         & non-local & full & channels & position & channel & weight \\
         \hline
         Local MLP & \cmark & \xmark & \cmark & \xmark & \cmark$^\flat$ & \xmark \\
         Local attention & \cmark & \xmark & \cmark & \xmark & \cmark$^\flat$ & \cmark \\
          DW-Conv. & \cmark & \xmark & \cmark & \cmark & \xmark & \xmark \\
          D-DW-Conv. & \cmark & \xmark & \cmark & \cmark & \xmark & \cmark \\
          Conv. & \cmark & \xmark & \xmark & \cmark  & \xmark & \xmark  \\
         \hline
         MLP & \xmark & \xmark & \xmark & \xmark  & \xmark & \xmark  \\
         Attention & \xmark & \xmark &\cmark & \xmark  & \cmark$^\flat$ & \cmark  \\
         Spatial-mixing MLP$^\dagger$  & \xmark & \xmark &  \cmark & \xmark  & \cmark & \xmark \\
         $1\times 1$ Conv.$^\ddagger$& \xmark & \cmark & \xmark & \cmark  & \xmark & \xmark  \\
         \shline
    \end{tabular}
    \vspace{-0.6cm}
\end{table}
 
\noindent\textbf{Convolution.}
Considering the $1$D case with a single channel
(the $2$D case is similar),
the connection weight matrix $\mathbf{W} \in {\mathbb{R}^{N \times N}}$
is in the following sparse form,
also known as the Toeplitz matrix
(We use the window size $3$ as an example):
\begin{align}
    \mathbf{W} =
\begin{bmatrix}
       a_2  & a_3 & 0 & 0 & \cdots & 0 & {a_1}           \\[0.3em]
       a_1  & a_2 & a_3 & 0 & \cdots & 0 & 0      \\[0.3em]
       \vdots & \vdots & \vdots & \vdots & \ddots    & \vdots
        & \vdots \\[0.3em]
       {a_3}  & 0 & 0 & 0 & \cdots & a_1 & a_2 
     \end{bmatrix}.
     \label{eqn:convsinglechannel}
\end{align}

For the $C$-channel case,
we organize the input into a vector channel by channel:
$[\mathbf{x}_1^\top~\mathbf{x}_2^\top~\dots ~\mathbf{x}_C^\top]^\top$,
and accordingly the connection weight matrix channel by channel
for the $c_o$th output channel, 
$\mathbf{W}_{c_o} = [\mathbf{W}_{c_o1}~\mathbf{W}_{c_o2}~\dots ~\mathbf{W}_{c_oC}]$
(the form of $\mathbf{W}_{c_oi}$ is 
the same as Equation~\ref{eqn:convsinglechannel}).
The whole form could be written as
\begin{align}
   \begin{bmatrix}
       \mathbf{y}_1\\[0.3em]
        \mathbf{y}_2 \\[0.3em]
         \vdots   \\[0.3em]
        \mathbf{y}_{{C }}
     \end{bmatrix} =
\begin{bmatrix}
       \mathbf{W}_1\\[0.3em]
        \mathbf{W}_2 \\[0.3em]
         \vdots   \\[0.3em]
        \mathbf{W}_{{C}}
     \end{bmatrix}
              \begin{bmatrix}
       \mathbf{x}_{1}\\[0.3em]
        \mathbf{x}_{2} \\[0.3em]
         \vdots   \\[0.3em]
        \mathbf{x}_{C}
     \end{bmatrix}.
\end{align}

\vspace{.1cm}
\noindent\textbf{Sep. MLP.}
Sep. MLP,
e.g., ResMLP and MLP-Mixer,
is formed with two kinds of 
block-sparse matrices:
one for channel-mixing
and the other for spatial-mixing.
In the case that the input is organized 
channel by channel
(the neurons in each channel form a group),
$\mathbf{x} = [\mathbf{x}_1^\top~\mathbf{x}_2^\top~\dots ~\mathbf{x}_C^\top]^\top$,
the connection weight matrix is in a block-sparse form:
\begin{align}
    \mathbf{W} =
\begin{bmatrix}
       \mathbf{W}_c    & \mathbf{0} & \cdots & \mathbf{0} & \mathbf{0}           \\[0.3em]
        \mathbf{0} & \mathbf{W}_c &  \cdots & \mathbf{0} & \mathbf{0}      \\[0.3em]
         \vdots & \vdots & \ddots    & \vdots
        & \vdots \\[0.3em]
        \mathbf{0} & \mathbf{0} & \cdots & \mathbf{0} & \mathbf{W}_c 
     \end{bmatrix},
\end{align}
where the block matrices $\mathbf{W}_c \in \mathbb{R}^{N\times N}$
are shared across all the channels,
and the sharing pattern can be modified 
to share weights within each group of channels.

The input can be reshaped 
position by position
(the neurons at each position forms a group):
$\mathbf{x} = [\mathbf{x}_1^\top~\mathbf{x}_2^\top~\dots ~\mathbf{x}_N^\top]^\top$,
and similarly one more connection weight matrix 
can be formulated in a block-sparse form
(it is essentially a $1\times 1$ convolution,
$\mathbf{W}_p \in \mathbb{R}^{C \times C}$):
\begin{align}
    \mathbf{W}' =
\begin{bmatrix}
       \mathbf{W}_p    & \mathbf{0} & \cdots & \mathbf{0} & \mathbf{0}           \\[0.3em]
        \mathbf{0} & \mathbf{W}_p &  \cdots & \mathbf{0} & \mathbf{0}      \\[0.3em]
         \vdots & \vdots & \ddots    & \vdots
        & \vdots \\[0.3em]
        \mathbf{0} & \mathbf{0} & \cdots & \mathbf{0} & \mathbf{W}_p 
     \end{bmatrix}.
\end{align}

The forms of block-sparsity
are studied in interleaved group convolutions~\citep{zhang2017interleaved}
without sharing the weights across groups.

Sep. MLP can also be regarded
as using Kronecker product to approximate the connection matrix, 
\begin{align}
\mathbf{W} \mathbf{x}
= \operatorname{vec}(\mathbf{A}\operatorname{mat}(\mathbf{x})\mathbf{B}).
\end{align}
Here, $\mathbf{W} = \mathbf{B}^\top \otimes \mathbf{A}
= \mathbf{W}_c^\top \otimes \mathbf{W}_p$.
and $\otimes$ is the Kronecker product operator.
$\operatorname{mat}(\mathbf{x})$
reshapes the vector $\mathbf{x}$
in a $2$D matrix form, while $\operatorname{vec}(\mathbf{x})$ reshapes the $2$D matrix into a vector form.
In Sep. MLP,
the $2$D matrix,
$\operatorname{mat}(\mathbf{x}) \in \mathbb{R}^{C\times N}$,
is organized 
so that each row corresponds to one channel
and each column corresponds to one spatial position.
CCNet~\citep{huang2019ccnet}
and interlaced self-attention~\citep{HuangYGZCW19}
use Kronecker product 
to approximate the spatial connection:
the former reshapes the vector in a $2$D matrix form
along the $x$ and $y$ axes, 
and the latter reshapes the vector windows by windows.

\vspace{.1cm}
\noindent\textbf{Vision Transformer (ViT).}
The matrix form is similar to Sep. MLP.
The difference is that
the matrix $\mathbf{W}_c$
is predicted from each image instance.
The weight prediction manner in ViT
has a benefit:
handle an arbitrary number of input neurons.

\vspace{.1cm}
\noindent\textbf{Depth-wise separable convolution.}
There are two basic components:
depth-wise convolution, and $1\times 1$ convolution that is the same as 
channel-mixing MLP in Sep. MLP.
Depth-wise convolution can be written in the matrix form:
\begin{align}
   \begin{bmatrix}
       \mathbf{y}_1\\[0.3em]
        \mathbf{y}_2 \\[0.3em]
         \vdots   \\[0.3em]
        \mathbf{y}_{C}
     \end{bmatrix} =
\begin{bmatrix}
       \mathbf{W}_{11} & \mathbf{0} & \cdots & \mathbf{0}\\[0.3em]
       \mathbf{0} & \mathbf{W}_{22} & \cdots & \mathbf{0}\\[0.3em]
       \vdots & \vdots & \ddots & \vdots\\[0.3em]
       \mathbf{0} & \mathbf{0} & \cdots & \mathbf{W}_{CC}\\[0.3em]
     \end{bmatrix}
       \begin{bmatrix}
       \mathbf{x}_{1}\\[0.3em]
        \mathbf{x}_{2} \\[0.3em]
         \vdots   \\[0.3em]
        \mathbf{x}_{C}
     \end{bmatrix},
\end{align}
where
the form of $\mathbf{W}_{cc}$ is 
the same as Equation~\ref{eqn:convsinglechannel}.

\vspace{.1cm}
\noindent\textbf{Local ViT.}
In the non-overlapping window partition case,
local ViT simply
repeats ViT over each window separately
with the linear projections, applied to
keys, values, and queries,
shared across windows.
In the overlapping case, the form is a little complicated, but the intuition is the same.
In the extreme case,
the partition is the same as convolution,
and the form is as the following:
\begin{align}
   \begin{bmatrix}
       \mathbf{y}_1\\[0.3em]
        \mathbf{y}_2 \\[0.3em]
         \vdots   \\[0.3em]
        \mathbf{y}_{C}
     \end{bmatrix} =
\begin{bmatrix}
       \mathbf{W}^d & \mathbf{0} & \cdots & \mathbf{0}\\[0.3em]
       \mathbf{0} & \mathbf{W}^d & \cdots & \mathbf{0}\\[0.3em]
       \vdots & \vdots & \ddots & \vdots\\[0.3em]
       \mathbf{0} & \mathbf{0} & \cdots & \mathbf{W}^d\\[0.3em]
     \end{bmatrix}
       \begin{bmatrix}
       \mathbf{x}_{1}\\[0.3em]
        \mathbf{x}_{2} \\[0.3em]
         \vdots   \\[0.3em]
        \mathbf{x}_{C}
     \end{bmatrix},
\end{align}
where the dynamic weight matrix 
$\mathbf{W}^d$ is like the form below:
\begin{align}
    \mathbf{W}^d =
\begin{bmatrix}
       a_{12}  & a_{13} & 0 & 0 & \cdots & 0 & {a_{11}}          \\[0.3em]
       a_{21}  & a_{22} & a_{23} & 0 & \cdots & 0 & 0      \\[0.3em]
       \vdots & \vdots & \vdots & \vdots & \ddots    & \vdots
        & \vdots \\[0.3em]
      {a_{N3}}    & 0 & 0 & 0 & \cdots & a_{N1} & a_{N2} 
     \end{bmatrix}.
\end{align}

\vspace{.1cm}
\noindent\textbf{Low-rank MLP.}
Low-rank MLP approximates the connection weight matrix $\mathbf{W} \in \mathbb{R}^{D_o \times D_i}$ in Equation~\ref{eqn:MLP}
using the product of two low-rank matrix:
\begin{align}
    \mathbf{W} \leftarrow 
    \mathbf{W}_{D_or}\mathbf{W}_{rD_i},
\end{align}
where $r$ is a number smaller than $D_i$ and $D_o$

\vspace{.1cm}
\noindent\textbf{Pyramid.}
The downsampling process in the pyramid networks
can be regarded as spatial low rank:
$\mathbf{W} (\in \mathbb{R}^{NC \times NC}) \rightarrow
\mathbf{W}' (\in \mathbb{R}^{N'C \times N'C})
$,
where $N'$ is equal to $\frac{N}{4}$
in the case that the resolution is reduced by $\frac{1}{2}$.
If the numbers of input and output channels are different,
it becomes
$\mathbf{W} (\in \mathbb{R}^{NC' \times NC}) \rightarrow
\mathbf{W}' (\in \mathbb{R}^{N'C' \times N'C})
$.

\vspace{.1cm}
\noindent\textbf{Multi-scale parallel convolution.}
Multi-scale parallel convolution used in HRNet~\citep{wang2020deep,SXLW19} can also be regarded as spatial low rank.
Consider the case with four scales,
multi-scale parallel convolution
can be formed as as the following,
\begin{align}
    \mathbf{W} \rightarrow
    \begin{bmatrix}
       \mathbf{W}_1 \in \mathbb{R}^{NC_1}\\[0.3em]
        \mathbf{W}_2 \in \mathbb{R}^{NC_2} \\[0.3em]
        \mathbf{W}_3 \in \mathbb{R}^{NC_3}\\[0.3em]
        \mathbf{W}_{4} \in \mathbb{R}^{NC_4}
     \end{bmatrix}
     \rightarrow 
     \begin{bmatrix}
       \mathbf{W}'_1 \in \mathbb{R}^{NC_1}\\[0.3em]
        \mathbf{W}'_2 \in \mathbb{R}^{\frac{N}{4}C_2} \\[0.3em]
        \mathbf{W}'_3 \in \mathbb{R}^{\frac{N}{16}C_3} \\[0.3em]
        \mathbf{W}'_{4} \in \mathbb{R}^{\frac{N}{64}C_4}
     \end{bmatrix},
\end{align}
where $C_1,C_2, C_3$, and $C_4$
are the numbers of the channels in four resolutions.

\section{Local Attention vs Convolution: Dynamic Weights}
\label{sec:dynamicweightcomparison}
We take the $1$D case with the window size $2K+1$ 
as an example
to illustrate 
the dynamic weight prediction manner.
Let $\{\mathbf{x}_{i-K}, \dots ,\mathbf{x}_{i}, \dots,
\mathbf{x}_{i+k}\}$ 
correspond to the $(2K+1)$ positions
in the $i$th window,
and 
$\{w_{i-K}, \dots, w_{i}, \dots, w_{i+K}\}$ 
be
the corresponding dynamic weights
for updating the representation of the $i$th (center) position.
The discussion can be easily extended
to multiple weights for each positions,
like the $M$-head attention
and updating the representations for other positions.

\noindent\textbf{Inhomogeneous dynamic convolution.}
We use the case using only a single linear projection
to illustrate inhomogeneous {dynamic convolution}.
The properties we will discuss
are similar for more linear projections.
The dynamic weights are predicted as the following:
\begin{align}
    \begin{bmatrix}
       w_{i-K}\\[0.3em]
       \vdots \\[0.3em]
        w_i\\[0.3em]
        \vdots \\[0.3em]
        w_{i+K}
     \end{bmatrix}
=\Theta \mathbf{x}_i 
    = \begin{bmatrix}
       \boldsymbol{\uptheta}_{-K}^\top\\[0.3em]
       \vdots \\[0.3em]
       \boldsymbol{\uptheta}_{0}^\top\\[0.3em]
       \vdots \\[0.3em]
        \boldsymbol{\uptheta}_{K}^\top
     \end{bmatrix}
    \mathbf{x}_i.
    \label{eqn:dynamicconvolutionweights}
\end{align}
It can be seen that
dynamic convolution learns the weights for each position 
through the parameters that are different for different positions,
e.g., $\bm{\uptheta}_{k}$
corresponds to $w_{i+k}$.
It regards the positions in the window
as the vector form, keeping the spatial order information.

\noindent\textbf{Dot-product attention.}
The~{dot-product attention} mechanism 
in the single-head case
predicts the weights as the following\footnote{For presentation clarity,
we omit the $\operatorname{softmax}$ normalization and the scale in dot-product.
What we discuss still holds 
if $\operatorname{softmax}$ and scale are included.}:
\begin{align}
    \begin{bmatrix}
       w_{i-K}\\[0.3em]
       \vdots \\[0.3em]
        w_i\\[0.3em]
        \vdots \\[0.3em]
        w_{i+K}
     \end{bmatrix}
    = 
    \begin{bmatrix}
       (\mathbf{x}_{i-K})^\top\\[0.3em]
       \vdots \\[0.3em]
        (\mathbf{x}_{i})^\top \\[0.3em]
        \vdots \\[0.3em]
        (\mathbf{x}_{i+K})^\top
     \end{bmatrix}
     \mathbf{P}_k^\top\mathbf{P}_q\mathbf{x}_i.
    \label{eqn:dotproductattentionweights}
\end{align}
Dot-product attention uses the same parameters $\mathbf{P}_k^\top\mathbf{P}_q$
for all the positions.
The weight depends on the features at the same position,
e.g.,
$w_{i-k}$
corresponds to $\mathbf{x}_{i-k}$.
It in some sense regards the positions in the window
as a set form, losing the spatial order information.

We rewrite it as the following
\begin{align}
\Theta_d = 
\begin{bmatrix}
       (\mathbf{x}_{i-K})^\top\\[0.3em]
       \vdots \\[0.3em]
        (\mathbf{x}_{i})^\top \\[0.3em]
        \vdots \\[0.3em]
        (\mathbf{x}_{i+K})^\top
     \end{bmatrix}
     \mathbf{P}_k^\top\mathbf{P}_q,
     \label{eqn:dot_attention}
\end{align}
from which we can see that 
the parameters $\Theta_d$
is dynamically predicted.
In other words, dot-product attention can be regarded
as a two-level dynamic scheme.

Relative position embeddings is equivalent
to adding static weights that keeps the spatial order information:
\begin{align}
    \begin{bmatrix}
       w_{i-K}\\[0.3em]
       \vdots \\[0.3em]
        w_i\\[0.3em]
        \vdots \\[0.3em]
        w_{i+K}
     \end{bmatrix} 
     = 
\Theta_d \mathbf{x}_i
     + \begin{bmatrix}
       \beta_{-K}\\[0.3em]
       \vdots \\[0.3em]
       \beta_{0} \\[0.3em]
       \vdots \\[0.3em]
        \beta_{K}
     \end{bmatrix}.
     \label{eqn:relativepositionembedding}
\end{align}

A straightforward variant is a combination
of the static $\Theta$
and the dynamic $\Theta_d$:
\begin{align}
        \begin{bmatrix}
       w_{i-K}\\[0.3em]
       \vdots \\[0.3em]
        w_i\\[0.3em]
        \vdots \\[0.3em]
        w_{i+K}
     \end{bmatrix} = 
(\Theta_d + \Theta) \mathbf{x}_i.
\end{align}
\noindent\textbf{Convolutional attention.}
We introduce a convolutional attention framework
so that it enjoys the benefits of dynamic convolution
and dot-product attention:
keep the spatial order information
and two-level dynamic weight prediction.

The post-convolutional attention mechanism
left-multiplies a matrix
(with the kernel size being $3$):
\begin{align}
\Theta_d 
     =
\begin{bmatrix}
       a_2  & a_3 & 0 & 0 & \cdots & 0 & {a_1}           \\[0.3em]
       a_1  & a_2 & a_3 & 0 & \cdots & 0 & 0      \\[0.3em]
       \vdots & \vdots & \vdots & \vdots & \ddots    & \vdots
        & \vdots \\[0.3em]
       {a_3}  & 0 & 0 & 0 & \cdots & a_1 & a_2 
     \end{bmatrix}
\begin{bmatrix}
       (\mathbf{x}_{i-K})^\top\\[0.3em]
       \vdots \\[0.3em]
        (\mathbf{x}_{i})^\top \\[0.3em]
        \vdots \\[0.3em]
        (\mathbf{x}_{i+K})^\top
     \end{bmatrix}
     \mathbf{P}_k^\top\mathbf{P}_q.
\end{align}
This can be reviewed as
a variant of relative positional embeddings
(Equation~\ref{eqn:relativepositionembedding}).
In the simplified case that the left matrix
is diagonal,
it can be regarded as the 
product version of relative positional embeddings
(Equation~\ref{eqn:relativepositionembedding}
is an addition version).

We can perform a convolution with the kernel size being $3$,
the kernel weights shared across channels 
(it is also fine not to share weights),
and then do dot-product attention.
This is called pre-convolutional attention:
perform convolutions on the representations.
The two processes
are can be written as follows
(omit BN and ReLU that follow the convolution),
\begin{align}
\begin{bmatrix}
       w_{i-K}\\[0.3em]
       \vdots \\[0.3em]
        w_i\\[0.3em]
        \vdots \\[0.3em]
        w_{i+K}
     \end{bmatrix}
     = \begin{bmatrix}
       a_1  & a_2 & a_3 &  \cdots & 0& 0 & 0      \\[0.3em]
       0  & a_1 & a_1 &  \cdots & 0& 0 & 0      \\[0.3em]
       \vdots & \vdots & \vdots & \ddots & \vdots    & \vdots
        & \vdots \\[0.3em]
        0 & 0 & 0 & \cdots  & a_2 & a_3 & 0 \\[0.3em]
         0 & 0 & 0 & \cdots & a_1 & a_2 & a_3
     \end{bmatrix}
\begin{bmatrix}
(\mathbf{x}_{i-K-1})^\top\\[0.3em]
       (\mathbf{x}_{i-K})^\top\\[0.3em]
       \vdots \\[0.3em]
        (\mathbf{x}_{i})^\top \\[0.3em]
        \vdots \\[0.3em]
        (\mathbf{x}_{i+K})^\top\\[0.3em]
        (\mathbf{x}_{i+K+1})^\top
     \end{bmatrix}
     \mathbf{P}_k^\top\mathbf{P}_q
     \begin{bmatrix}
     \mathbf{x}_{i-1} & \mathbf{x}_i & \mathbf{x}_{i+1}
     \end{bmatrix}
     \begin{bmatrix}
       a_1 \\[0.3em]
       a_2 \\[0.3em]
       a_3
     \end{bmatrix}.
\end{align}
It can be generalized to using normal convolution:
\begin{align}
\begin{bmatrix}
       w_{i-K}\\[0.3em]
       \vdots \\[0.3em]
        w_i\\[0.3em]
        \vdots \\[0.3em]
        w_{i+K}
     \end{bmatrix}
     = \mathbf{C}'
\begin{bmatrix}
       \mathbf{x}_{i-K-1} & \mathbf{x}_{i-K-1} & \cdots & \mathbf{x}_{i-K-1}\\[0.3em]
       \mathbf{x}_{i-K} & \mathbf{x}_{i-K} & \cdots & \mathbf{x}_{i-K}\\[0.3em]
       \vdots & \vdots & \ddots & \vdots\\[0.3em]
        \mathbf{x}_{i} & \mathbf{x}_{i} & \cdots & \mathbf{x}_{i}\\[0.3em]
        \vdots & \vdots & \ddots & \vdots\\[0.3em]
        \mathbf{x}_{i+K} & \mathbf{x}_{i+K} & \cdots & \mathbf{x}_{i+K} \\[0.3em]
        \mathbf{x}_{i+K+1} & \mathbf{x}_{i+K+1} & \cdots & \mathbf{x}_{i+K+1} 
     \end{bmatrix}
     \mathbf{P}_k^\top\mathbf{P}_q
     \mathbf{C}_3
     \begin{bmatrix}
     \mathbf{x}_{i-1} \\[0.3em]
     \mathbf{x}_i \\[0.3em]
     \mathbf{x}_{i+1}
     \end{bmatrix}.
\end{align}
Here,
$\mathbf{C}$' is a ($2K+1$)-row matrix and
can be easily derived
from the convolutional kernel
$\mathbf{C}_3$.
The $(2K+1)$ weights, 
$\{w_{i-1}, w_{i},
w_{i+1}\}$,
correspond to the $(2K+1)$ rows in $\mathbf{C}$, respectively.
This means that
the three positions
are differentiated
and the same position in each window
corresponds to the same row. 
This explains why the positional embeddings
are not necessary when convolutions are adopted~\citep{wu2021cvt}.
Using different pairs $(\mathbf{W}_q, \mathbf{W}_k)$
leads to more weights for each position,
e.g., $M$ pairs correspond to $M$-head attention.

\section{Architecture Details}
\label{sec:architecturedetails}
\noindent\textbf{Overall structures.}
Following local vision transformer,
Swin Transformer~\citep{liu2021swin}, 
we build two depth-wise convolution-based
networks, namely DWNet-T and DWNet-B. 
The corresponding dynamic versions are
dynamic DWNet-T, dynamic DWNet-B, i-dynamic DWNet-T, 
and i-dynamic DWNet-B.
The depth-wise convolution-based networks
follow the overall structure of Swin Transformer. 
We replace local self attention by depth-wise convolution with the same 
window size.
We use batch normalization~\citep{ioffe2015batch} and 
ReLU~\citep{nair2010rectified} instead of layer
normalization~\citep{ba2016layer}
in the convolution blocks.

Table~\ref{tab:arch-spec} shows the architecture details
of Swin Transformer and DWNet for the tiny model.
Normalizations are performed within the residual 
block, same as Swin Transformer.
The base model is similarly built
by following Swin Transformer
to change the number of channels
and the depth of the third stage.

\begin{table}[t!]
\footnotesize
\centering
\addtolength{\tabcolsep}{3pt}
\renewcommand{\arraystretch}{1.4}
\caption{Architectures details of Swin Transformer and DWNet for the tiny model. The architectures for the base model
can be easily obtained.}
\begin{tabular}{c|c|c|c}
\shline
 & \makecell{downsp. rate \\ (output size)} & Swin  &  DWNet \\
\hline 
\multirow{6}{*}{stage 1} & \multirow{6}{*}{\makecell{4$\times$\\ (56$\times$56)}} & concat 4$\times$4, linear 96-d, LN  & concat 4$\times$4, linear 96-d, LN  \\
\cline{3-4}
& & $\begin{bmatrix}
\text{LN, linear 96x3-d}\\\text{local sa. 7$\times$7, head 3}\\\text{linear 96-d}
\\ 
\text{LN, linear 384-d}\\\text{GELU, linear 96-d} 
\end{bmatrix}$ $\times$ 2   
& $\begin{bmatrix}\text{linear 96-d, BN, ReLU}\\\text{depthwise conv. 7$\times$7, BN, ReLU}\\\text{linear 96-d, BN} \\

\text{BN, linear 384-d} \\
\text{GELU, linear 96-d}
\end{bmatrix}$ $\times$ 2 \\
\hline
\multirow{6}{*}{stage 2}  & \multirow{6}{*}{\makecell{8$\times$\\ (28$\times$28)}} & concat 2$\times$2, linear 192-d , LN & concat 2$\times$2, linear 192-d , LN \\
\cline{3-4}
& & $\begin{bmatrix}
\text{LN, linear 192x3-d}\\\text{local sa. 7$\times$7, head 6}\\\text{linear 192-d}
\\ 
\text{LN, linear 768-d}\\\text{GELU, linear 192-d} 
\end{bmatrix}$ $\times$ 2   
& $\begin{bmatrix}\text{linear 192-d, BN, ReLU}\\\text{depthwise conv. 7$\times$7, BN, ReLU}\\\text{linear 192-d, BN} \\

\text{BN, linear 768-d} \\
\text{GELU, linear 192-d}
\end{bmatrix}$ $\times$ 2 \\
\hline
\multirow{6}{*}{stage 3}  & \multirow{6}{*}{\makecell{16$\times$\\ (14$\times$14)}}  & concat 2$\times$2, linear 384-d , LN & concat 2$\times$2, linear 384-d , LN  \\
\cline{3-4}
& & $\begin{bmatrix}
\text{LN, linear 384x3-d}\\\text{local sa. 7$\times$7, head 12}\\\text{linear 384-d}
\\ 
\text{LN, linear 1536-d}\\\text{GELU, linear 384-d} 
\end{bmatrix}$ $\times$ 6   
& $\begin{bmatrix}\text{linear 384-d, BN, ReLU}\\\text{depthwise conv. 7$\times$7, BN, ReLU}\\\text{linear 384-d, BN} \\

\text{BN, linear 1536-d} \\
\text{GELU, linear 384-d}
\end{bmatrix}$ $\times$ 6 \\
\hline
\multirow{6}{*}{stage 4} & \multirow{6}{*}{\makecell{32$\times$\\ (7$\times$7)}}  & concat 2$\times$2, linear 768-d , LN & concat 2$\times$2, linear 768-d , LN \\
\cline{3-4}
& & $\begin{bmatrix}
\text{LN, linear 768x3-d}\\\text{local sa. 7$\times$7, head 24}\\\text{linear 768-d}
\\ 
\text{LN, linear 3072-d}\\\text{GELU, linear 768-d} 
\end{bmatrix}$ $\times$ 2   
& $\begin{bmatrix}\text{linear 768-d, BN, ReLU}\\\text{depthwise conv. 7$\times$7, BN, ReLU}\\\text{linear 768-d, BN} \\

\text{BN, linear 3072-d} \\
\text{GELU, linear 768-d}
\end{bmatrix}$ $\times$ 2 \\
\hline 
\multirow{2}{*}{stage 4} & \multirow{2}{*}{\makecell{1$\times$1}} & 
LN, AvgPool. 1$\times$1 & LN, AvgPool. 1$\times$1  \\
& & linear classifier & linear classifier \\
\shline
\end{tabular}
\label{tab:arch-spec}
\end{table}

\noindent\textbf{Dynamic depth-wise
convolution.} 
Dynamic depth-wise convolution generates 
the connection weights according to the 
instance.
As described in Section~\ref{sec:dynamicdw},
for the homogeneous version, 
we conduct the global average pooling operation
to get a vector,
and adopt two linear projections:
the first one reduces the dimension by $1/4$,
followed by BN and ReLU,
and then generate the kernel weights and shared 
for all spatial positions.
Unlike SENet~\citep{hu2018squeeze},
we currently do not use
the Sigmoid activation function
for generating the weights.
For the inhomogeneous version, we generate unshared
dynamic weight
for each spatial position using the corresponding feature.
The connection weights are shared across channels to reduce 
the model parameters and computation complexity. 
Specifically, we share $3$ and $4$ channels in each group
of channels for tiny and base models. Thus the number
of model parameters and computation complexity are similar
to Swin Transformer.

\begin{table}[t!]
        \centering
            \footnotesize
    \setlength{\tabcolsep}{3pt}
            \renewcommand{\arraystretch}{1.14}
            
        \caption{\small ImageNet classification comparison
        for ResNet, HRNet,
        Mixer and ResMLP and gMLP,
        ViT and DeiT,
        Swin (Swin Transformer),
        DWNet,
        dynamic DWNet and i-dynamic DWNet.
        \dag~means
        that ResNet is built
        by 
        using two $3 \times 3$ convolutions 
        to form the residual units.
        Table~\ref{tab:regularizationcomparisonappendix}
        presents the comparison 
        for representative modules
        in terms of spare connectivity,
        weight sharing and dynamic weight.
        }
        \label{tab:imagenetclassification1}
        \begin{tabular}{l | c r r c | c c }
        \shline 
        method & \makecell{img. size} & \#param. & FLOPs &  \makecell{throughput (img. / s)} & \makecell{top-1 acc.}&  \makecell{real acc.}\\
        \hline 
        \multicolumn{7}{l}{\emph{Convolution: local connection}}\\
        \hline
        ResNet-38~\dag~\citep{wang2020deep} &   224$^2$ & 28M & 3.8G & 2123.7 & 75.4 & - \\
        ResNet-72~\dag~\citep{wang2020deep} &  224$^2$ &  48M & 7.5G & 623.0 &  76.7 & - \\ 
        ResNet-106~\dag~\citep{wang2020deep} &  224$^2$ & 65M & 11.1G & 452.8 & 77.3 & - \\
        \hline
        \multicolumn{7}{l}{\emph{Bottleneck:
        convolution with low rank}}\\
        \hline
        ResNet-50~\citep{he2016deep} & 224$^2$ & 26M & 4.1G & 1128.3 & 76.2 & 82.5 \\
        ResNet-101~\citep{he2016deep} &  224$^2$ & 45M & 7.9G & 652.0 &  77.4 & 83.7 \\ 
        ResNet-152~\citep{he2016deep} & 224$^2$ &60M & 11.6G & 456.7 & 78.3 & 84.1 \\
        \hline 
        \multicolumn{7}{l}{\emph{Pyramid:
        convolution with pyramid (spatial low rank) features. }}\\
        \hline
        HRNet-W18~\citep{wang2020deep} & 224$^2$ & 21M & 4.0G & - & 76.8 & - \\
        HRNet-W32~\citep{wang2020deep} & 224$^2$ & 41M & 8.3G & - & 78.5 & - \\
        HRNet-W48~\citep{wang2020deep} & 224$^2$ & 78M & 16.1G & - & 79.3 & - \\
        \hline
        \multicolumn{7}{l}{\emph{Channel and spatial separable MLP, spatial separable MLP = point-wise $1\times 1$ convolution}}\\
        \hline
        Mixer-B/16~\citep{tolstikhin2021mlp} & 224$^2$ &46M & - & - & 76.4 & 82.4\\ 
        Mixer-L/16~\citep{tolstikhin2021mlp} & 224$^2$ &189M & - & - & 71.8 & 77.1\\
        ResMLP-12~\citep{touvron2021resmlp} & 224$^2$ &15M & 3.0G & - & 76.6 & 83.3  \\ 
        ResMLP-24~\citep{touvron2021resmlp} & 224$^2$ &30M & 6.0G & - & 79.4 & 85.3\\ 
        ResMLP-36~\citep{touvron2021resmlp} & 224$^2$ &45M & 8.9G & - & 79.7 & 85.6\\ 
        gMLP-Ti~\citep{liu2021pay} & 224$^2$ & 6M & 1.4G & - & 72.0 & - \\
        gMLP-S~\citep{liu2021pay} & 224$^2$ & 20M & 4.5G & - & 79.4 & - \\
        gMLP-B~\citep{liu2021pay} & 224$^2$ & 73M & 15.8G & - & 81.6 & - \\
        \hline 
        \multicolumn{7}{l}{\emph{Global attention: dynamic channel separable MLP + spatial separable MLP}}\\
        \hline
        ViT-B/16~\citep{dosovitskiy2021an} & 384$^2$ &86M & 55.4G & 83.4 & 77.9 & 83.6 \\ 
        ViT-L/16~\citep{dosovitskiy2021an} & 384$^2$ &307M & 190.7G & 26.5 & 76.5 & 82.2 \\
        DeiT-S~\citep{touvron2020training} & 224$^2$ &22M & 4.6G  & 947.3 & 79.8 & 85.7\\
        DeiT-B~\citep{touvron2020training} & 224$^2$ &86M & 17.5G & 298.2 & 81.8 & 86.7\\
        DeiT-B~\citep{touvron2020training} & 384$^2$ &86M & 55.4G & 82.7 & 83.1 & 87.7\\
        \hline 
        \multicolumn{7}{l}{\emph{Pyramid attention: perform attention
        with spatial low rank}}\\
        \hline
        PVT-S~\citep{wang2021pyramid} & 224$^2$ & 25M & 3.8G & - & 79.8 & - \\
        PVT-M~\citep{wang2021pyramid} & 224$^2$ & 44M & 6.7G & - & 81.2 & - \\
        PVT-L~\citep{wang2021pyramid} & 224$^2$ & 61M & 9.8G & - & 81.7 & - \\
        \hline
        \multicolumn{7}{l}{\emph{Local MLP: perform static separable MLP
        in local small windows}}\\
        \hline
        Swin-Local MLP-T & 224$^2$&  26M & 3.8G & 861.0 & 80.3 & 86.1\\
        Swin-Local MLP-B & 224$^2$&  79M & 12.9G & 321.2 & 82.2 & 86.9 \\
        \hline
        \multicolumn{7}{l}{\emph{Local attention: perform attention
        in local small windows}}\\
        \hline
        Swin-T~\citep{liu2021swin} & 224$^2$&28M & 4.5G & 713.5 & 81.3 & 86.6\\
        Swin-B~\citep{liu2021swin} &224$^2$&  88M & 15.4G & 263.0 & 83.3 & 87.9\\
        \hline
        \multicolumn{7}{l}{\emph{Depth-wise convolution + point-wise $1\times 1$ convolution}}\\
        \hline
        DWNet-T &  224$^2$&24M & 3.8G & 928.7 & 81.3 & 86.8 \\
        DWNet-B & 224$^2$& 74M & 12.9G & 327.6 & 83.2 & 87.9\\
        dynamic DWNet-T & 224$^2$ & 51M & 3.8G & 897.0 & 81.9 & 87.3\\
        dynamic DWNet-B & 224$^2$ & 162M & 13.0G & 322.4 & 83.2 & 87.9\\
        i-dynamic DWNet-T & 224$^2$ & 26M & 4.4G & 685.3 & 81.8 & 87.1\\
        i-dynamic DWNet-B & 224$^2$ & 80M & 14.3G & 244.9 & 83.4 & 88.0\\
        \shline
        \end{tabular}
        \vspace{-0.5cm}
\end{table}

\section{Setting Details}
\label{sec:settingdetails}
\noindent\textbf{ImageNet pretraining.}
We use the identical training setting with Swin Transformer
in ImageNet pretraining for fair comparison.
The default input size is $224 \times 224$. 
The AdamW optimizer~\citep{loshchilov2017decoupled}, 
with the initial 
learning rate $0.001$
and
the weight decay $0.05$, is used for 300 epochs. 
The learning rate is scheduled by a cosine decay schema and warm-up with linear schema for the first 20 epochs.
We train the model on 8 GPUs with the total batch size 1024.
The augmentation and regularization strategies are same as 
Swin Transformer, which includes RandAugment~\citep{cubuk2020randaugment},
Mixup~\citep{zhang2018mixup}, CutMix~\citep{yun2019cutmix}, 
random erasing~\citep{zhong2020random} 
and stochastic depth~\citep{huang2016deep}. 
The stochastic depth rate
is employed as $0.2$ and $0.5$ for 
the tiny and base models, respectively,
the same as Swin Transformer. 

\noindent\textbf{COCO object detection.}
We follow Swin Transformer to adopt Cascade Mask
R-CNN~\citep{cai2019cascade} for comparing backbones.
We use the training and test settings from Swin Transformer: multi-scale training - resizing the
input such that the shorter side is between 480 and 800 and the longer side is at most 1333;
AdamW optimizer with the initial learning rate
0.0001; 
weight decay - 0.05;
batch size - 16;
and epochs - 36.

\noindent\textbf{ADE semantic segmentation.}
Following Swin Transformer,
we use UPerNet~\citep{xiao2018unified} as the segmentation framework.
We use the same setting as the Swin Transformer:
the AdamW optimizer with initial learning rate 0.00006;
weight decay 0.01;
linear learning rate decay;
160,000 iterations with warm-up for 1500 iterations;
8 GPUs with mini-batch 2 per GPU.
We use the same data augmentation as 
Swin Transformer based on MMSegmentation~\citep{mmseg2020}.
The experimental results are reported as single scale testing.

\noindent\textbf{Static version of Swin Transformer - Local MLP.}
We remove the linear projections applied 
to keys and queries, 
accordingly dot production and softmax normalization. 
The connection weights 
(corresponding to attention weights
in the dynamic version) are set as static model 
parameters which are learnt during the training
and shared for all the images.

\noindent\textbf{Retraining on $384 \times 384$.}
We retrain the depth-wise convolution-based network on the ImageNet dataset with
$384 \times 384$ input images 
from the model trained with $224 \times 224$ images.
We use learning rate 
$10^{-5}$, weight decay 
$10^{-8}$ and stochastic depth ratio $0.1$ for $30$ epochs
for both $7\times 7$
and $12 \times 12$ windows.

\section{Additional Experiments and Analysis} 
\label{sec:experiments}

\noindent\textbf{More results on ImageNet classification.}
We give more experimental results 
with different sparse connection strategies, as shown in
Table~\ref{tab:imagenetclassification1}. These results
also verify that locality-based sparsity pattern 
(adopted in depth-wise convolution and local attention) 
besides sparsity between channels/spatial positions still facilitates the network training for ImageNet-1K.

\noindent\textbf{Results on large scale pre-training.}
Transformers~\citep{liu2021swin,dosovitskiy2021an} 
show higher performance compared with 
the previous convolutional networks with 
large scale pre-training.
We further study the performance on ImageNet-22K pre-training.
We first train the model on ImageNet-22K dataset
which has about 14.2 million images, and then
fine-tune the model on ImageNet-1K classification, downstream detection and segmentation tasks.
The same training settings with Swin transformer are used in all tasks.
The fine-tuning results
in Table~\ref{tab:imagenet22k} and Table~\ref{tab:coco_det_ade_seg_22k}  
indicate the dynamic convolution based DWNets 
could get the performance 
comparable to Swin transformer 
with large scale pre-training.
\begin{table}[b]
    \centering
    \footnotesize
    \setlength{\tabcolsep}{12pt}
    \renewcommand{\arraystretch}{1.15}
    \caption{ Comparison on ImageNet-1K classification with ImageNet-22K pre-training.
    }
    \label{tab:imagenet22k}
    \begin{tabular}{l | r r c}%
    \shline
      & \multicolumn{3}{c}{ImageNet-1K fine-tuning}  \\ 
      \cline{2-4}
      & \#param. & FLOPs & top-1 acc. \\ 
      \hline
      Swin-B & 88M & 15.4G & 85.2 \\ 
      DWNet-B & 74M & 12.9G & 84.8 \\
      dynamic DWNet-B & 162M & 13.0G & 85.0 \\
      i-dynamic DWNet-B & 80M & 14.3G & 85.2 \\
     \shline
    \end{tabular}
\end{table}

\begin{table}[t]
        \centering
            \footnotesize
    \setlength{\tabcolsep}{4pt}
            \renewcommand{\arraystretch}{1.15}
        \caption{Comparison results on COCO object detection and ADE semantic segmentation with ImageNet-22k pre-training.}
        \label{tab:coco_det_ade_seg_22k}
        \vspace{-0.3cm}
        \begin{tabular}{l | r c| c  c c c  | r r | c}
        \shline
        \multirow{2}{*}{}& \multicolumn{6}{c|}{{COCO fine-tuning}} & \multicolumn{3}{c}{{ADE20K fine-tuning}} \\ 
        \cline{2-10}
        
         & \#param. & FLOPs &  AP$^{box}$ & AP$^{box}_{50}$ & AP$^{box}_{75}$ & AP$^{mask}$  & \#param. & FLOPs  &  mIoU  \\
        
        \hline 
        Swin-B & 145M & 986G & 53.4 & 72.1 & 58.1 & 46.1 & 121M & 1192G & 49.4  \\ 
        DWNet-B & 132M & 924G & 52.0 & 70.4 & 56.3 & 45.0 & 108M & 1129G & 50.1 \\ 
        dynamic DWNet-B & 219M & 924G & 51.9 & 70.7 & 56.2 & 45.0 & 195M & 1129G &  49.6 \\
        i-dynamic DWNet-B & 137M & 948G & 52.9 & 71.2 & 57.2 & 45.8 & 114M & 1153G &  51.3 \\
        \shline
        \end{tabular}
        \vspace{-.1cm}
\end{table}

\noindent\textbf{Cooperating with different normalization functions.}
Transformers usually use the layer
normalization to stabilize the training, while
convolutional architectures adopt batch normalization.
We verify different combinations of backbones (Swin and DWNet) and normalization functions. The popular 
used layer normalization (LN), batch normalization (BN),
and the dynamic version of batch normalization - centering
calibrated batch normalization~\citep{gao2021representative}
(CC. BN) are verified 
in the experiments.
Table~\ref{tab:normalization} shows the results
on ImageNet classification.

\begin{table}[t]
           \centering
            \footnotesize
    \setlength{\tabcolsep}{10pt}
            \renewcommand{\arraystretch}{1.15}
        \caption{Exploring normalization schemes
        of Swin Transformer and depth-wise convolution
        based networks (DWNet) for the tiny model.
        The results are reported on the ImageNet top-1 accuracy.}
        \label{tab:normalization}
\begin{tabular}{l|c c c c  c c  }
        \shline
            &  Layer Norm. &  Batch Norm. &  Centering calibrated Batch Norm. & Top-1 Acc.  \\
            \hline
            Swin & \cmark & &  & $81.3$\\
            Swin &  & \cmark &  & $80.9$\\
            Swin &  & & \cmark & $81.2$\\
            \hline
            DWNet & \cmark &  &  & $81.2$ \\
            DWNet & & \cmark &  & $81.3$ \\
            DWNet & & & \cmark & $81.7$ \\
                
        \shline
        \end{tabular}
\end{table}

\begin{table}[t]
    \centering
    \footnotesize
    \setlength{\tabcolsep}{5pt}
    \renewcommand{\arraystretch}{1.15}
    \caption{ Comparison between local attention
    and depth-wise convolution in
    VOLO~\citep{yuan2021volo} and SVT~\citep{chu2021twins}
    architecture. Results are reported on ImageNet classification with tiny model.
    }
    \label{tab:volo}
    \begin{tabular}{l | c c c c }
    \shline
      & \#param. & FLOPs & top-1 acc.\\
      \hline
      VOLO-d1~\citep{yuan2021volo} & 27M & 7.0G & 84.1 \\ 
      VOLO (Local SA)-d1 & 27M & 7.2G & 84.2 \\
      DW Conv.-d1 & 26M & 6.9G & 84.2 \\ 
      \hline 
      SVT-S~\citep{chu2021twins} & 24M & 2.8G & 81.7 \\
      DW Conv.-S & 22M & 2.7G & 81.9 \\
     \shline
    \end{tabular}
\end{table}

\noindent\textbf{Depth-wise convolution with other architectures.}
We conduct experiments on other local attention designs,
such as SVT~\citep{chu2021twins} and
VOLO~\citep{yuan2021volo} whose implementations are
publicly available. SVT uses local self attention as a
basic spatial feature fusion operation, while VOLO
proposes a new attention module named Vision Outlooker.
We replace the local self attention with depth-wise
convolution in SVT same as the paper, and replace Vision
Outlooker with $7\times7$ local self attention and 
$7\times7$ depth-wise
convolution, respectively. The remaining structures are
unchanged and the same training setting is used as the
original papers.
The experimental results are shown in Tab~\ref{tab:volo}
and the observations are the same as the 
Swin Transformer design.

\noindent\textbf{Retraining on $384\times 384$ images.}
Similar to~\citep{liu2021swin},
we study the performance of fine-tuning
the models:
first learn with $224\times 224$ images,
then fine-tune on large images of $384\times 384$.
We study two cases:
(1) keep the window size $7\times 7$ unchanged;
and (2) upsample the kernel weights from $7\times 7$ to $12\times 12$
as done in~\citep{liu2021swin} for upsampling the relative positional embeddings.

The results are in Table~\ref{tab:finetune}\footnote{Swin Transformer takes slightly higher FLOPs for $7\times 7$ than $12 \times 12$. The higher computation cost comes
from larger padding than $12\times 12$.}.
In the case of keeping the window size $7\times 7$ unchanged,
DWNet performs better.
When using a larger window size $12\times 12$, 
depth-wise convolution performs worse than $7\times 7$.
We suspect that this is because upsampling the kernel weights 
is not a good starting for fine-tuning.
In Swin Transformer,
using a larger window size improves the performance.
We believe that this is because the local attention mechanism
is suitable for variable window sizes.

\noindent\textbf{Cooperating with SE.}
Squeeze-and-excitation~\citep{hu2018squeeze} (SE)
is a parameter- and computation-efficient dynamic module,
initially designed for improving the ResNet performance.
The results in Table~\ref{tab:se} show that 
DWNet, a static module, benefits
from the SE module,
while Swin Transformer,
already a dynamic module,
does not benefit from dynamic module SE.
The reason is still unclear,
and might lie in the optimization.

\begin{table}[t!]
\makeatletter\def\@captype{table}
\centering
\setlength{\tabcolsep}{3pt}
\footnotesize
\renewcommand{\arraystretch}{1.1}
\caption{\footnotesize {Retrain on larger images.}}
\label{tab:finetune}
\vspace{-0.2cm}
\footnotesize
\begin{tabular}{c | c c c c }
            \shline
                model & \makecell{ws.} & \#param. & FLOPs & Acc. \\
                \hline
                \multirow{2}{*}{Swin} & 7$\times$7 & 28M & 14.4G & 81.8 \\
                & 12$\times$12 & 28M & 14.2G & 82.4 \\
                \hline
                \multirow{2}{*}{DWNet}   & 7$\times$7 & 24M & 11.1G & 82.2 \\
                 & 12$\times$12 & 25M & 11.5G & 82.1 \\
            \shline
\end{tabular}
\vspace{-0.3cm}
\end{table}

\begin{table}[t!]
\centering
\setlength{\tabcolsep}{4pt}
\footnotesize
\renewcommand{\arraystretch}{1.1}
\caption{\footnotesize{Cooperate with SE.}}
\vspace{-0.2cm}
\label{tab:se}
\footnotesize
\begin{tabular}{c | c c c c}
            \shline
                model & SE & \#param. & FLOPs & Acc. \\
                \hline
                \multirow{2}{*}{Swin} & \xmark & 28M & 4.5G & 81.3  \\
                & \cmark & 29M & 4.5G & 81.2 \\
                \hline
                \multirow{2}{*}{DWNet}  & \xmark & 24M & 3.8G & 81.3 \\
                 & \cmark & 24M & 3.8G & 81.7 \\
            \shline
\end{tabular}
\vspace{-0.3cm}
\end{table}

\section{Potential Studies}

\noindent\textbf{Complexity balance
between point-wise ($1\times1$) convolution and depth-wise (spatial) convolution.}
Depth-wise convolution takes only about $2\%$ computation in the depth-wise convolution-based architecture.
The major computation complexity comes from $1\times 1$ convolutions.
The solutions to this issue could be:
group $1\times1$ convolution studied in IGC~\citep{zhang2017interleaved,sun2018igcv3},
and channel-wise weighting (like SENet) 
studied in Lite-HRNet~\citep{yu2021lite}
and EfficientNet~\citep{tan2019efficientnet,tan2021efficientnetv2},
or simply add more depth-wise (spatial) convolutions.

\noindent\textbf{Attention weights as channel maps.}
Attention weights in attention 
can be regarded as channel maps.
The operations, such as convolution or simple weighting,
can be applied to the attention weights.
The resT approach~\citep{zhang2021rest}
performs $1\times 1$ convolutions over
the attention weight maps.

\noindent\textbf{Dynamic weights.}
In Swin Transformer
and our developed dynamic depth-wise convolution networks,
only the spatial part, attention and depth-wise convolution,
explores dynamic weights.
Lite-HRNet instead studies dynamic weight for point-wise ($1\times 1$) convolution.
It is interesting to explore dynamic weight for both parts.

\noindent\textbf{Convolution-style MLP weights.}
The weights of the spatial-mixing MLP in MLP-Mixer and ResMLP
could be modified
in the convolution-like style
with more weights
(some like the relative position embeddings used in local attention, larger than the image window size)
so that it could be extended to larger images and downstream tasks with different image sizes.

\end{document}